\documentclass[twoside,11pt]{article}

\usepackage[preprint]{jmlr2e}

\usepackage{multirow}
\usepackage{amsmath}
\usepackage{amsfonts}
\usepackage{subfig}
\usepackage{caption}
\usepackage{algorithm}
\usepackage{algorithmic}
\usepackage{wrapfig}
\usepackage{multicol}
\usepackage{enumitem}
\usepackage[dvipsnames]{xcolor}
\usepackage{resizegather}

\ShortHeadings{Graph Representation Learning for Multi-Task Settings}{Buffelli and Vandin}
\firstpageno{1}

\begin{document}

\title{Graph Representation Learning for Multi-Task Settings: a Meta-Learning Approach}

\author{\name Davide Buffelli \email davide.buffelli@unipd.it \\
       \addr Department of Information Engineering\\
       University of Padova\\
       Padova, Italy
       \AND
       \name Fabio Vandin \email fabio.vandin@unipd.it \\
       \addr Department of Information Engineering\\
       University of Padova\\
       Padova, Italy}

\maketitle

\begin{abstract}
Graph Neural Networks (GNNs) have become the state-of-the-art method for many applications on graph structured data. 
GNNs are a model for \textit{graph representation learning}, which aims at learning to generate low dimensional node embeddings that encapsulate structural and feature-related information.
GNNs are usually trained in an end-to-end fashion, leading to highly specialized node embeddings. While this approach achieves great results in the single-task setting, the generation of node embeddings that can be used to perform multiple tasks (with performance comparable to single-task models) is still an open problem.
We propose the use of meta-learning to allow the training of a GNN model capable of producing \textit{multi-task} node embeddings. 
In particular, we exploit the properties of optimization-based meta-learning to learn GNNs that can produce general node representations by learning parameters that can quickly (i.e. with a few steps of gradient descent) adapt to multiple tasks.
Our experiments show that the embeddings produced by a model trained with our purposely designed meta-learning procedure can be used to perform multiple tasks with comparable or, surprisingly, even higher performance than both single-task and multi-task end-to-end models.
\end{abstract}


\section{Introduction}\label{intro}
Graph Neural Networks (GNNs) are deep learning models for graph structured data, and have become one of the main topics of the deep learning research community. The interest in GNNs is due, in part, to their great empirical performance on many graph-related tasks. Three tasks in particular, with many practical applications, have received the most attention: graph classification, node classification, and link prediction.

\begin{figure}[h]
\centering
  \subfloat{\includegraphics[width=0.75\linewidth]{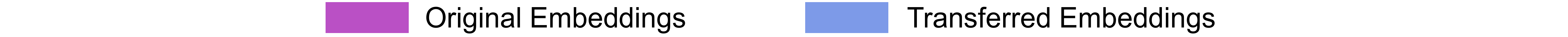}}\\
  \vspace{-2mm}
  \subfloat[      (a)]{\includegraphics[width=0.2\linewidth]{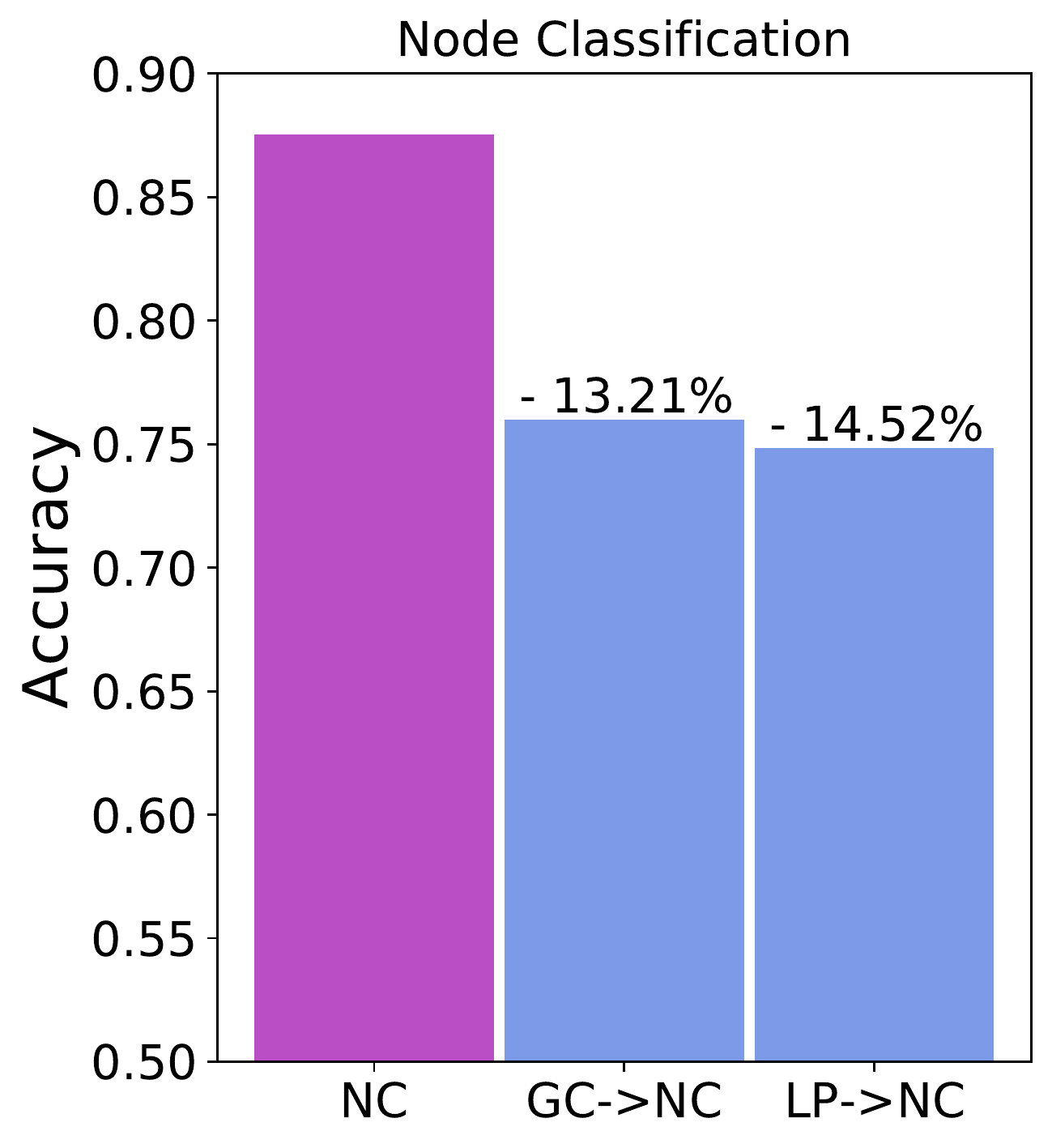}}
  \subfloat[      (b)]{\includegraphics[width=0.2\linewidth]{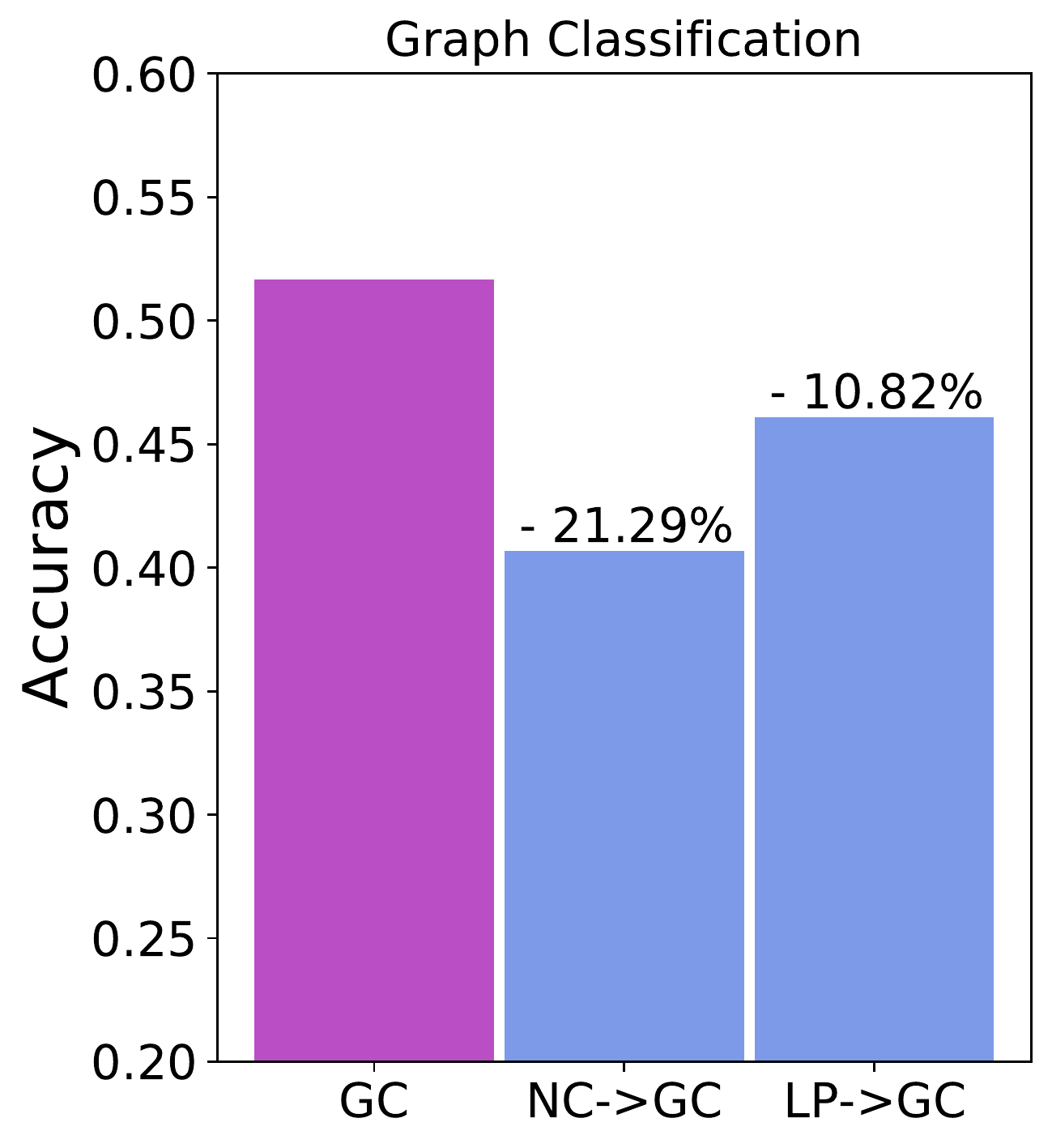}}
  \subfloat[      (c)]{\includegraphics[width=0.2\linewidth]{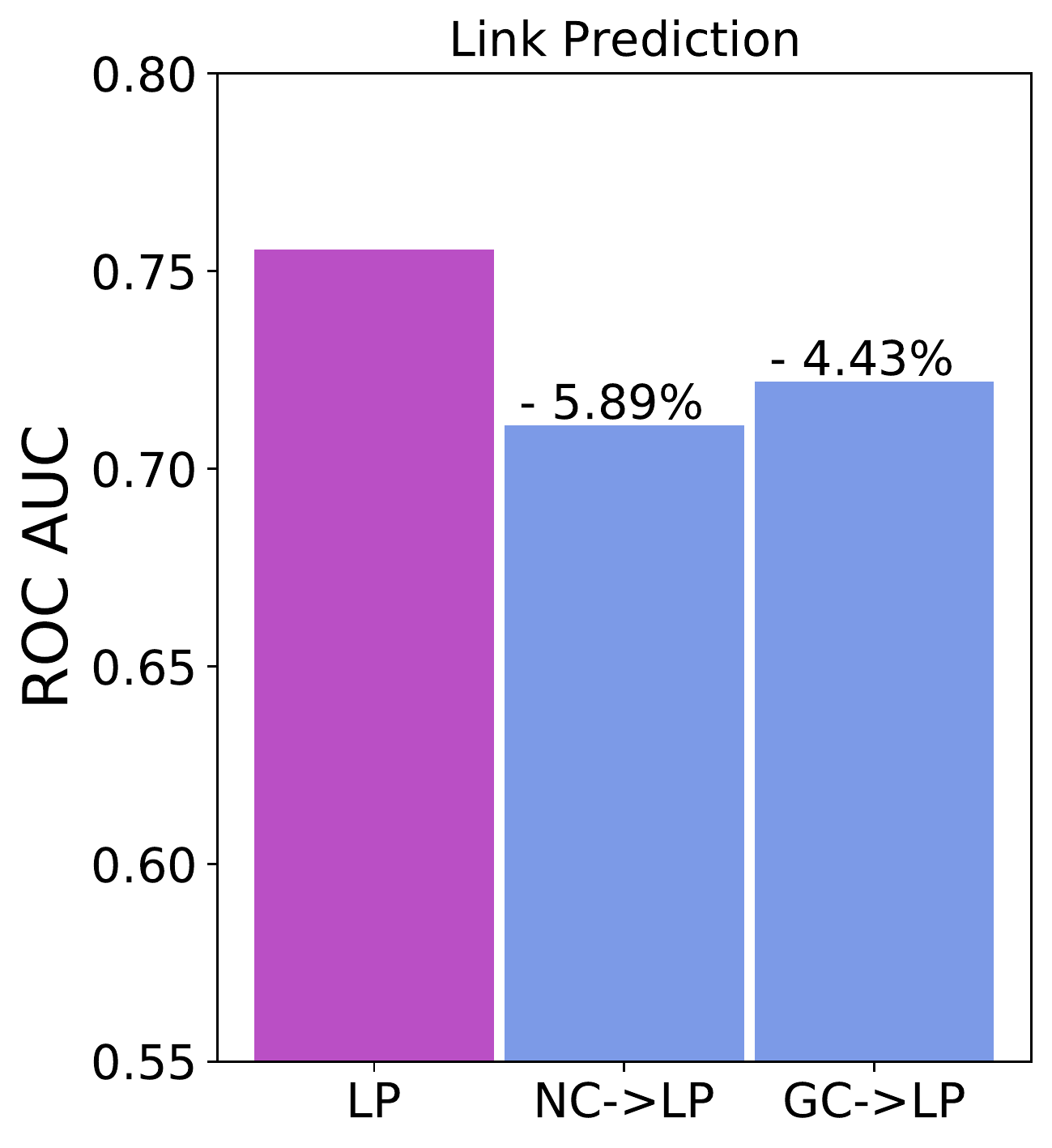}}
  \caption{Performance drop when transferring node embeddings on the ENZYMES dataset to perform the tasks: (a) Node Classification (NC), (b) Graph Classification (GC), and (c) Link Prediction (LP). 
  ``\textit{x -\textgreater y}'' indicates that the embeddings obtained from a model trained on task \textit{x} are used to train a network for task \textit{y}.}
  \label{fig:transfer_embeddings}
\end{figure}

GNNs are centered around the concept of \textit{node representation learning}, and typically follow the same architectural pattern with an \textit{encoder-decoder} structure \citep{hamilton2017representation,Chami2020MachineLO,Wu_2020}. The encoder produces node embeddings (low-dimensional vectors capturing relevant structural and feature-related information about each node), while the decoder uses the embeddings to carry out the desired downstream task. The model is then trained in an end-to-end manner, leading to highly specialized node embeddings. While this approach can achieve state-of-the-art performance, it also affects the generality and reusability of the embeddings. In fact, taking the node embeddings generated by an encoder trained for a given task, and using them to train a decoder for a different task leads to substantial performance loss (see Fig. \ref{fig:transfer_embeddings}).

The low transferability of node embeddings requires the use of one specialized encoder and one specialized decoder for each considered task. However, there are many practical scenarios in which \emph{multiple} tasks must be performed on the same graph(s). For example, in a social network, both classification tasks (e.g., classify users as spammer or nonspammer) and link prediction tasks (e.g., suggest new connections between users) must be performed. While a trivial solution might be to deploy a specific model (generating specific node embeddings) for each task, this inevitably leads to significant overhead. 

\looseness=-1
This paper studies the problem of generating node embeddings that can be used for multiple tasks, which is a scenario with many practical applications that, nonetheless, has received little attention from the GNN community. In more detail, we propose a multi-task representation learning procedure, based on optimization-based meta-learning \citep{finn2017}, that learns a GNN encoder producing node embeddings that generalize across multiple tasks. Our focus is on the most studied tasks in the GNN literature: graph classification, node classification, and link prediction.



The proposed meta-learning procedure is targeted towards multi-task representation learning and takes advantage of MAML \citep{finn2017} and ANIL \citep{raghu2019rapid} to reach a setting of the parameters where a few steps of gradient descent on a given task lead to good performance on that task. This procedure leads to an encoder-decoder model that can easily be adapted to perform each of the tasks \emph{singularly}, and hence encourages the encoder to learn representations that can be reused across tasks. At the end of the training procedure, the decoder is discarded, and the encoder GNN is used to generate embeddings.


We summarize our contributions as follows:
\begin{itemize}

\item We consider the under-studied problem of learning GNN models generating node representations that can be used to perform multiple tasks. In this regard, we design a meta-learning strategy for training GNN models with such capabilities. 

\item To the best of our knowledge, we are the first to propose a GNN model generating a \textit{single} set of node embeddings that can be used to perform the three most common graph-related tasks (i.e., graph classification, node classification, and link prediction). In particular, the generated embeddings lead to comparable or even higher performance with respect to separate end-to-end trained single-task models. 
\item We show that the episodic training strategy at the base of our meta-learning procedure leads to a model generating node embeddings that are more effective for downstream tasks, even in single-task settings. This unexpected finding is of interest in itself, and may provide fruitful directions for future research.
\end{itemize}

\section{Preliminaries}
This section introduces GNNs (Section~\ref{sec:prelims_GNNs}), multi-head models (Section~\ref{multihead}), and optimization-based meta-learning techniques (Section~\ref{maml_anil}), which are at the core of our method. Throughout the paper we use the term ``\textit{task}'' as in the multi-task learning (MTL) literature, i.e. to refer to a downstream application (e.g. graph classification, node classification, etc.).

\subsection{Graph Neural Networks}
\label{sec:prelims_GNNs}
Many popular state-of-the-art GNN models follow the \textit{message-passing} paradigm \citep{10.5555/3305381.3305512}, which we now briefly describe. We represent a graph $\mathcal{G} = (\mathbf{A}, \mathbf{X})$ with an adjacency matrix $\mathbf{A} \in \{0,1\}^{n \times n}$, such that $\mathbf{A}_{ij}=1$ if and only if there is an edge between the $i$-th vertex and the $j$-th vertex, and a node feature matrix $\mathbf{X} \in \mathbb{R}^{n \times d}$, where the $v$-th row $\mathbf{X}_v$ represents the $d$ dimensional feature vector of node $v$. Let $\mathbf{H}^{(\ell)} \in \mathbb{R}^{n \times d^{\prime}}$ be the matrix containing the node representations at layer $\ell$. A message passing layer updates the representation of every node $v$ as follows:
\begin{align*}
\text{\textit{msg}}^{(\ell)}_{v}  &= \text{AGGREGATE}(\{ \mathbf{H}_{u}^{(\ell)} \text{ } \forall u \in \mathcal{N}_v \}) \\
\mathbf{H}_{v}^{(\ell+1)} &= \text{UPDATE}(\mathbf{H}_{v}^{(\ell)}, \text{\textit{msg}}^{(\ell)}_{v})   
\end{align*}
where $\mathbf{H}^{(0)} = \mathbf{X}$, $\mathcal{N}_v$ is the set of neighbours of node $v$, $\text{AGGREGATE}$ is a permutation invariant function (as it takes a set as input), and $\text{UPDATE}$ is usually a neural network. After $L$ message-passing layers, the final node embeddings $\mathbf{H}^{(L)}$ are the representations used to perform a given task (e.g., they are the input to a neural component that performs the given task), and the network is trained end-to-end. 

\subsection{Multi-Head Models}\label{multihead}

\begin{figure*}[t]
\centering
  \subfloat[(a)]{\includegraphics[width=0.20\textwidth]{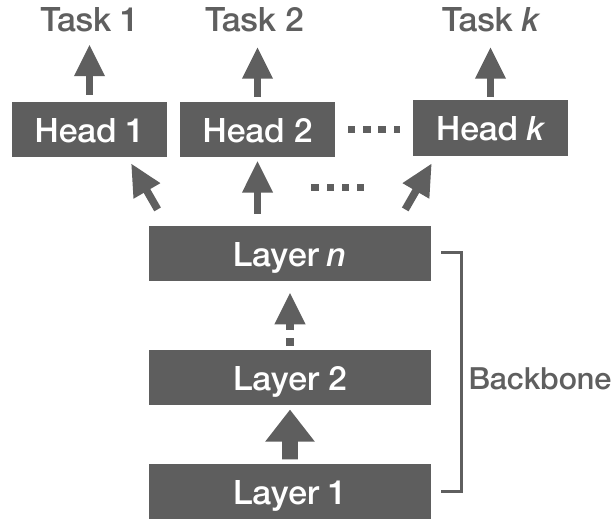}}\hspace{\fill} \vline \hspace{\fill}
  \subfloat[(b)]{\includegraphics[width=0.20\textwidth]{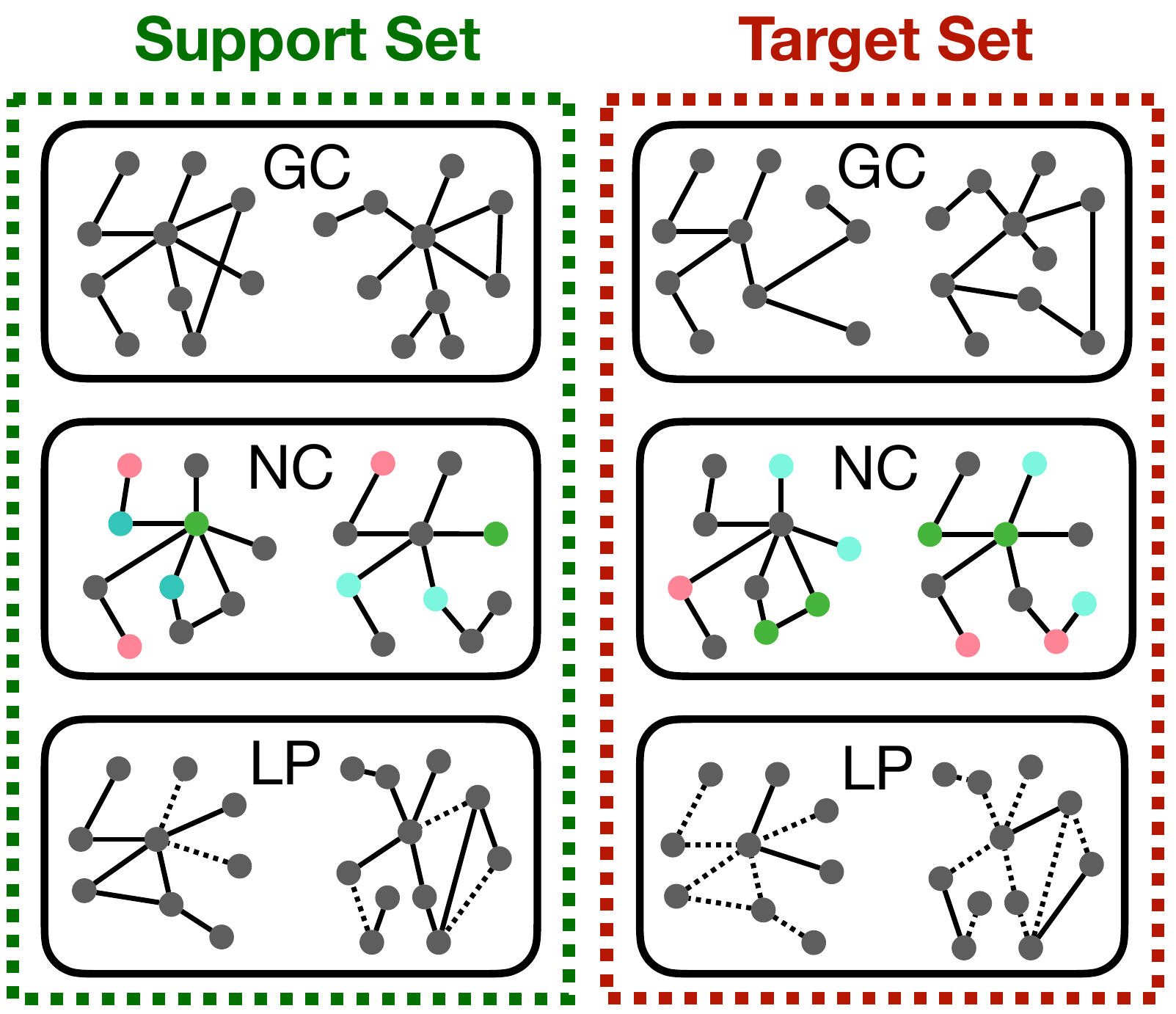}}\hspace{\fill}
  \subfloat[(c)]{\includegraphics[height=1in]{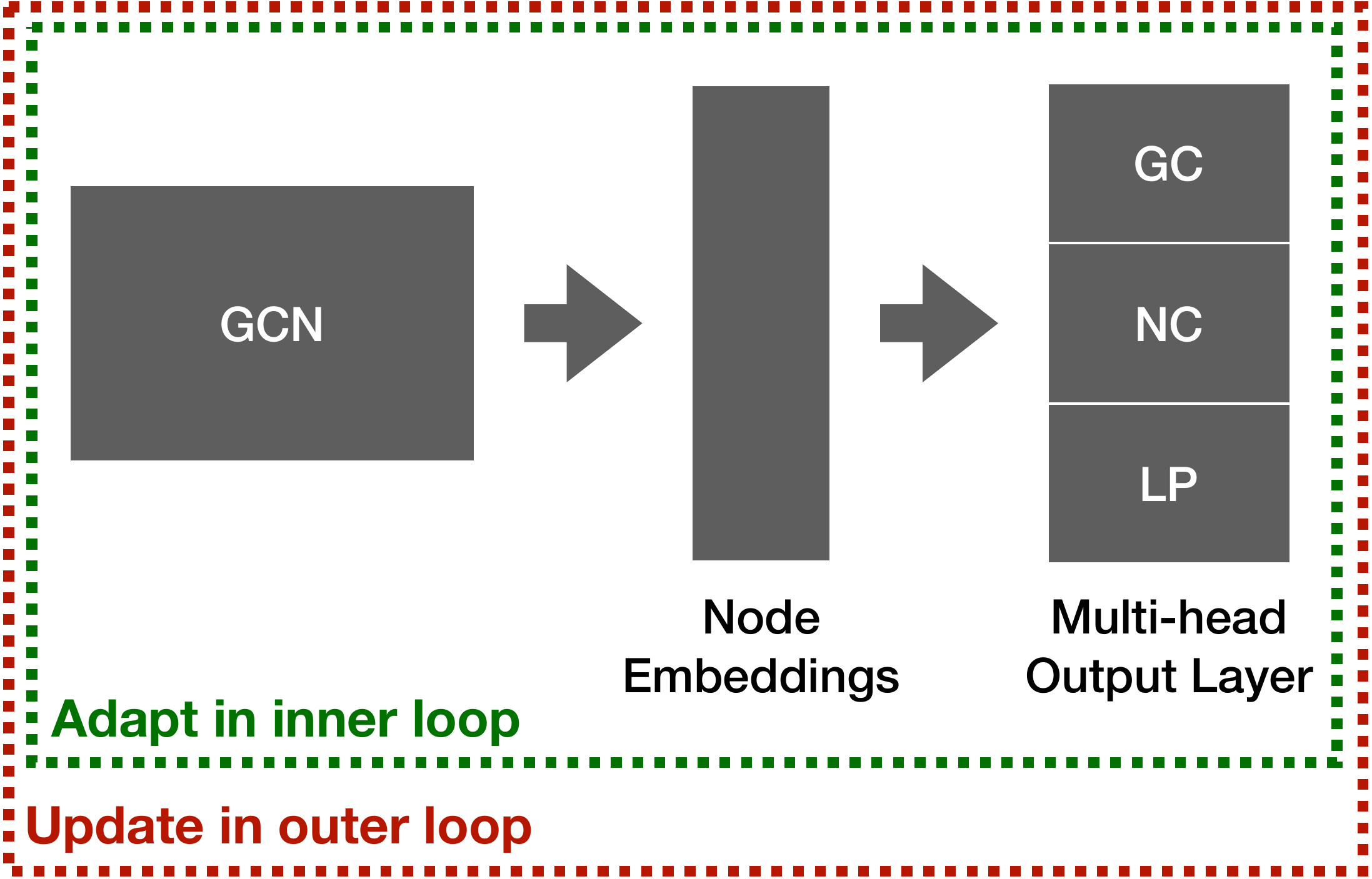}}\hspace{\fill}
  \subfloat[(d)]{\includegraphics[height=1in]{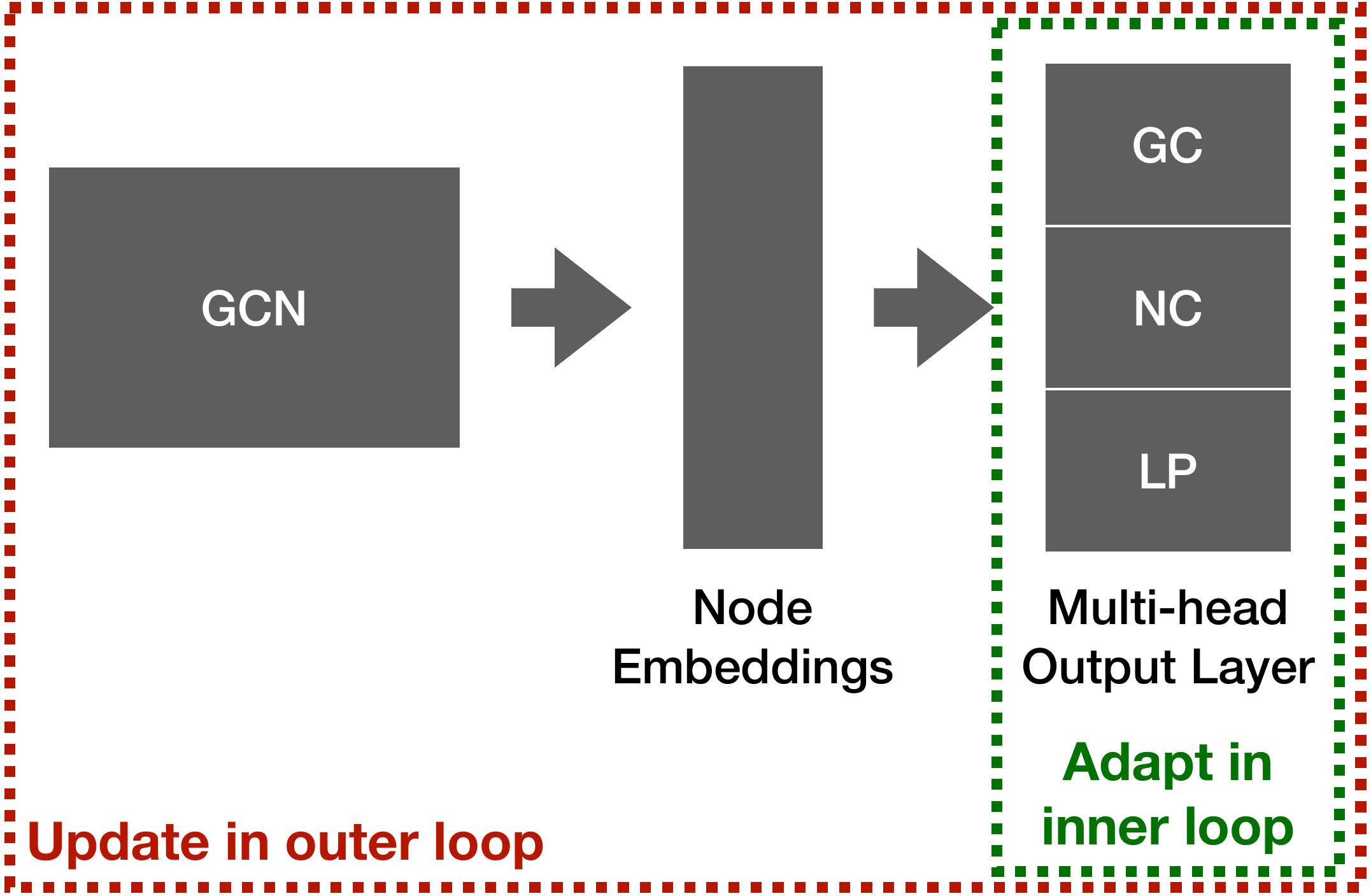}}
  \caption{\textbf{Main ingredients of our meta-learning procedure SAME.} (a) Multi-head architecture. (b) Schematic representation of a \textit{multi-task episode}. For each task, support and target set are designed to be as the training and validation sets for single-task training. (c-d) Overview of the parameter updates in SAME's meta-learning procedure. In the inner loop, the model is adapted \textit{separately} to each task in the support set; the outer loop then tests the performance of each adaptation on the corresponding task in the target set, and updates the initial parameters of the network to allow it to rapidly be adapted to each task, by minimizing the meta-objective. In iSAME (c) all parameters are adapted in the inner loop. In eSAME (d) only the task-specific output layers are adapted in the inner loop. Both in iSAME and eSAME, after training the model, only the backbone GCN is kept, and used to generate embeddings.}
  \label{fig:models}
\end{figure*}

In deep learning, the standard approach for performing multiple tasks \citep{v2020revisiting} with the same model is to employ a \textit{multi-head} architecture (see Fig. \ref{fig:models} (a)). A multi-head model is composed of a \textit{backbone} and multiple \textit{heads} (one for each task). The backbone is a neural network which processes the input to extract features. The features extracted by the backbone are then used by the heads (which are also neural networks) to perform the desired tasks (each head performs one task). The whole model is then trained end-to-end to minimize a combination of the single-task losses (e.g. the sum of the losses on each task). We refer to this strategy as the \textit{classical} training procedure for multi-task models.

\subsection{Model-Agnostic Meta-Learning and ANIL}\label{maml_anil}

MAML (Model-Agnostic Meta-Learning) is an optimization-based meta-learning strategy proposed by \citet{finn2017}. Let $f_\theta$ be a deep learning model, where $\theta$ represents its parameters. Let $p(\mathcal{E})$ be a distribution over episodes\footnote{The meta-learning literature usually derives episodes from \textit{tasks} (i.e., tuples containing a dataset and a loss function). We focus on episodes to avoid using the term \textit{task} for both a MTL task, and a meta-learning task.}, with an episode $\mathcal{E}_i \sim p(\mathcal{E})$ being defined as a tuple containing a \textit{loss function} $\mathcal{L}_{\mathcal{E}_i}(\cdot)$, a \textit{support set} $\mathcal{S}_{\mathcal{E}_i}$, and a \textit{target set} $\mathcal{T}_{\mathcal{E}_i}$: $\mathcal{E}_i = (\mathcal{L}_{\mathcal{E}_i}(\cdot), \mathcal{S}_{\mathcal{E}_i}, \mathcal{T}_{\mathcal{E}_i})$, where support and target sets are simply sets of labelled examples. MAML's goal is to find a value of $\theta$ that can quickly, i.e. in a few steps of gradient descent, be adapted to new episodes. This is done with a nested loop optimization procedure: an \textit{inner loop} adapts the parameters to the support set of an episode by performing some steps of gradient descent, and an \textit{outer loop} updates the initial parameters aiming at a setting that allows fast adaptation. Formally, by defining $\theta^{\prime}_{i} (t)$ as the parameters after $t$ adaptation steps on the support set of episode $\mathcal{E}_i$, we can express the computations in the inner loop as
\begin{equation}\label{eq_inner}
 \theta^{\prime}_{i} (t) = \theta^{\prime}_{i} (t-1) - \alpha \nabla_{\theta^{\prime}_{i} (t-1)} \mathcal{L}_{\mathcal{E}_i}(f_{\theta^{\prime}_{i} (t-1)}, \mathcal{S}_{\mathcal{E}_i})
\end{equation}
where $\theta^{\prime}_{i} (0) = \theta$, $\mathcal{L}(f_{\theta^{\prime}_{i} (t-1)}, \mathcal{S}_{\mathcal{E}_i})$ indicates the loss over the support set $\mathcal{S}_{\mathcal{E}_i}$ for the model $f_{\theta^{\prime}_{i} (t-1)}$ with parameters $\theta^{\prime}_{i} (t-1)$, and $\alpha$ is the learning rate. The \textit{meta-objective} that the outer loop tries to minimize is defined as
$ \mathcal{L}_{\text{\textit{meta}}} = \sum_{\mathcal{E}_{i} \sim p(\mathcal{E})} \mathcal{L}_{\mathcal{E}_i}(f_{\theta^{\prime}_{i} (t)}, \mathcal{T}_{\mathcal{E}_i}) $,
which leads to the following parameter update\footnote{We limit ourself to one step of gradient descent for clarity, but any optimization strategy could be used.} performed in the outer loop:
\begin{equation}
\theta = \theta - \beta \nabla_{\theta}\mathcal{L}_{\text{\textit{meta}}} = \theta - \beta \nabla_{\theta} \sum_{\mathcal{E}_{i} \sim p(\mathcal{E})} \mathcal{L}_{\mathcal{E}_i}(f_{\theta^{\prime}_{i} (t)}, \mathcal{T}_{\mathcal{E}_i}).
\end{equation}

\citet{raghu2019rapid} showed that feature reuse is the dominant factor in MAML: in the adaptation loop, only the last layer(s) in the network are updated, while the first layer(s) remain almost unchanged. The authors then propose ANIL (Almost No Inner Loop) where they split the parameters in two sets: one that is used for adaptation in the inner loop, and one that is only updated in the outer loop. This simplification leads to computational improvements while maintaining performance.

\section{SAME: \underline{S}ingle-Task \underline{A}daptation for \underline{M}ulti-Task \underline{E}mbeddings}
We design a meta-learning approach targeted towards representation learning, by building on three insights: 

\textbf{\textit{(i)} optimization-based meta-learning is implicitly learning robust representations.} The findings by \citet{raghu2019rapid} suggest that, in a model trained with MAML, the first layers learn features that are reusable across episodes, while the last layers are set up for fast adaptation. MAML is then \textit{implicitly} learning a model with two components: an \textit{encoder} (the first layers), focusing on learning reusable representations that generalize across episodes, and a \textit{decoder} (the last layers) that can be quickly adapted for different episodes. 

\textbf{\textit{(ii)} meta-learning episodes can be designed to encourage generalization.} By designing support and target sets to mimic the training and validation sets of a classical training procedure, then the meta-learning procedure is effectively optimizing for generalization.

\textbf{\textit{(iii)} meta-learning can learn to quickly adapt to multiple tasks \textit{singularly}, without having to learn to solve multiple tasks \textit{concurrently}.} 
The meta-learning procedure can be deisgned so that, for each considered task, the inner loop adapts the parameters to a task-specific support set, and tests the adaptation on a task-specific target set. The outer loop then updates the parameters to allow this fast \textit{multiple single-task adaptation}.

Based on \textbf{\textit{(ii})} and \textbf{\textit{(iii})}, we design the meta-learning procedure such that the inner loop adapts to multiple tasks \textit{singularly}, each time with the goal of \textit{single-task generalization}. Using an encoder-decoder architecture, \textbf{\textit{(i})} suggests that this procedure leads to an encoder that learns features reusable across episodes. As, in each episode, the learner is adapting to multiple tasks, the encoder is learning features that generalize across multiple tasks. After training with our meta-learning strategy, the decoder is discarded, and only the encoder is kept and used to generate representations. Contrary to many applications of meta-learning, there is no adaptation performed at test time, as meta-learning is used only for training the model from which an encoder is exracted.


In the rest of this section, we formally present our meta-learning procedure for training multi-task graph representation learning models. There are three aspects that need to be defined: \textbf{(1) Episode Design:} how is a an episode composed, \textbf{(2) Model Architecture Design:} what is the architecture of our model, \textbf{(3) Meta-Training Design:} how, and which, parameters are adapted/updated.

\subsection{Episode Design}\label{epdesign} 
In our case, an episode becomes a \textit{multi-task episode} (Fig. \ref{fig:models} (b)). To formally introduce the concept, let us consider the case where the tasks are graph classification (GC), node classification (NC), and link prediction (LP). We define a \textit{multi-task episode} $\mathcal{E}^{(m)}_{i} \sim p(\mathcal{E}^{(m)})$ as a tuple $\mathcal{E}^{(m)}_{i} = ( \mathcal{L}^{(m)}_{\mathcal{E}_{i}},  \mathcal{S}^{(m)}_{\mathcal{E}_{i}},   \mathcal{T}^{(m)}_{\mathcal{E}_{i}} )$ where
\begin{align*}
\mathcal{L}^{(m)}_{\mathcal{E}_{i}} &= \{ \mathcal{L}^{(\text{GC})}_{\mathcal{E}_{i}}, \mathcal{L}^{(\text{NC})}_{\mathcal{E}_{i}}, \mathcal{L}^{(\text{LP})}_{\mathcal{E}_{i}} \}, \\
\mathcal{S}^{(m)}_{\mathcal{E}_{i}} &= \{ \mathcal{S}^{(\text{GC})}_{\mathcal{E}_i}, \mathcal{S}^{(\text{NC})}_{\mathcal{E}_i}, \mathcal{S}^{(\text{LP})}_{\mathcal{E}_i} \}, \\
\mathcal{T}^{(m)}_{\mathcal{E}_{i}} &= \{ \mathcal{T}^{(\text{GC})}_{\mathcal{E}_i}, \mathcal{T}^{(\text{NC})}_{\mathcal{E}_i}, \mathcal{T}^{(\text{LP})}_{\mathcal{E}_i} \}.
\end{align*}
The meta-objective $\mathcal{L}^{(m)}_{\text{\textit{meta}}}$ of our method is then defined as:
 \begin{gather}
\label{eq:meta_obj}
 \mathcal{L}^{(m)}_{\text{\textit{meta}}}=  \sum_{\mathcal{E}^{(m)}_{i} \sim p(\mathcal{E}^{(m)})} \lambda^{(GC)} \mathcal{L}^{(\text{GC})}_{\mathcal{E}_{i}} + \lambda^{(NC)} \mathcal{L}^{(\text{NC})}_{\mathcal{E}_{i}} + \lambda^{(LP)} \mathcal{L}^{(\text{LP})}_{\mathcal{E}_{i}}.
\end{gather}
where $\lambda^{(\cdot)}$ are balancing coefficients.
 
Support and target sets are set up to resemble training and validation sets. This way the outer loop's objective becomes to \textit{maximize the performance on a validation set, given a training set}, hence encouraging generalization.
In more detail, given a batch of graphs, we divide it in equally sized splits (one per task), and create support and target sets as follows:
\begin{description}
\item \textbf{Graph Classification:} $\mathcal{S}^{(\text{GC})}_{\mathcal{E}_i}$ and $\mathcal{T}^{(\text{GC})}_{\mathcal{E}_i}$ contain labeled graphs, obtained with a random split.
\item \textbf{Node Classification:} $\mathcal{S}^{(\text{NC})}_{\mathcal{E}_i}$ and $\mathcal{T}^{(\text{NC})}_{\mathcal{E}_i}$ are composed of the same graphs, with different labelled nodes.
We mimic the common semi-supervised setting \citep{kipf2017semi} where feature vectors are available for all nodes, and only a small subset of nodes is labelled.
\item \textbf{Link Prediction:} $\mathcal{S}^{(\text{LP})}_{\mathcal{E}_i}$ and $\mathcal{T}^{(\text{LP})}_{\mathcal{E}_i}$ are composed of the same graphs, with different query edges. In every graph we randomly remove some edges, used as positive examples together with non-removed edges, and randomly sample pairs of non-adjacent nodes as negative examples.
\end{description}
Notice how we only need labels for \textit{one} task for each graph. The full algorithm for the creation of \textit{multi-task episodes} is provided in Appendix\footnote{Appendix  available at: \url{https://arxiv.org/abs/2201.03326}.} A. 

\subsection{Model Architecture Design}\label{architecture}
We use an encoder-decoder model with a multi-head architecture. The \textit{backbone} (which represents the encoder) is composed of 3 GCN \citep{kipf2017semi} layers with ReLU non-linearities and residual connections \citep{7780459}. The decoder is composed of three \textit{heads}. The node classification head is a single layer neural network with a \textit{Softmax} activation that is shared across nodes and maps node embeddings to class predictions. In the graph classification head, first a single layer neural network (shared across nodes) performs a linear transformation (followed by a ReLU activation) of the node embeddings. The transformed node embeddings are then averaged and a final single layer neural network with \textit{Softmax} activation outputs the class predictions. The link prediction head is composed of a single layer neural network with ReLU non-linearity that transforms node embeddings, and a single layer neural network that given concatenation of two embeddings outputs the probability of a link between them. We remark that after training the full model with the proposed meta-learning procedure, only the encoder is kept, and is used to generate node embeddings which can be fed to any machine learning model for downstream tasks.

\subsection{Meta-Training Design}\label{adaptations}
We first present the meta-learning training procedure, and successively describe which parameters are adapted/updated in the inner and outer loops.

\textbf{Meta-Learning Training Procedure.}
The meta-learning procedure is designed such that the inner loop adaptation 
involves a \textit{single task} at a time. Only the parameter update performed to minimize the meta-objective involves multiple tasks, but,  crucially, it does not aim at a setting of parameters that can solve, or quickly adapt to, multiple tasks \textit{concurrently}, but to a setting allowing \textbf{multiple fast single-task adaptation}. 

  \begin{algorithm}[tb]
   \caption{Proposed (meta-learning based) procedure.}\label{meta_alg}
   \begin{algorithmic}
       \STATE {\bfseries Input:} Model $f_{\theta}$; Episodes $\mathcal{E} = \{ \mathcal{E}_1, .., \mathcal{E}_n \}$; Meta-Objective Coefficients $\lambda^{(GC)}, \lambda^{(NC)}, \lambda^{(LP)}$. \;
      \vspace{0.5mm}
      \STATE \texttt{init}$(\theta)$  \; 
      \FOR{$\mathcal{E}_i$  {\bfseries in} $\mathcal{E}$}
         \STATE $\text{\texttt{o\_loss}} \leftarrow 0$ \;
         
         \FOR{\text{\texttt{$\tau$}} {\bfseries in} (GC, NC, LP)}
            \STATE $\theta^{\prime (\text{\texttt{$\tau$}})} \leftarrow \theta$ \;
            \STATE $\theta^{\prime (\text{\texttt{$\tau$}})} \leftarrow \text{\texttt{ADAPT}}(f_{\theta}, \mathcal{S}^{(\texttt{$\tau$})}_{\mathcal{E}_i}, \mathcal{L}^{(\texttt{$\tau$})}_{\mathcal{E}_i})$ \;
            \STATE $\text{\texttt{o\_loss}} \leftarrow \text{\texttt{o\_loss}} + \lambda^{(\tau)} \text{\texttt{TEST}}(f_{\theta^{\prime (\text{\texttt{$\tau$}})}}, \mathcal{T}^{(\texttt{$\tau$})}_{\mathcal{E}_i}, \mathcal{L}^{(\texttt{$\tau$})}_{\mathcal{E}_i})$
           \ENDFOR
           
         \STATE $\theta \leftarrow \text{\texttt{UPDATE}}(\theta, \text{\texttt{o\_loss}}, \theta^{\prime (GC)}, \theta^{\prime (NC)}, \theta^{\prime (LP)})$
       \ENDFOR     
      \end{algorithmic}
   \end{algorithm}
   

The pseudocode of our procedure is in Algorithm \ref{meta_alg}. \texttt{init} is a method that initializes the weights of the GNN. \texttt{ADAPT} performs a few steps of gradient descent on a \textit{task-specific} loss function and support set (as in eq. \ref{eq_inner}), \texttt{TEST} computes the value of the meta-objective component on a \textit{task-specific} loss function and target set for a model with parameters adapted on that task, and \texttt{UPDATE} optimizes the parameters $\theta$ by minimizing the meta-objective in eq. ~\ref{eq:meta_obj} (which is contained in \texttt{o\_loss} in the pseudocode). Notice the multiple heads of the decoder are never used concurrently.

\textbf{Parameter Update in Inner/Outer Loop.}
Let us partition the parameters of our model in four sets: $\theta = [ \theta_{\text{GCN}}, \theta_{\text{NC}}, \theta_{\text{GC}}, \theta_{\text{LP}} ]$ representing the parameters of the backbone ($\theta_{GCN}$), node classification head ($\theta_{NC}$), graph classification head ($\theta_{GC}$), and link prediction head ($\theta_{LP}$). We name our meta-learning strategy SAME (\underline{S}ingle-Task \underline{A}daptation for \underline{M}ulti-Task \underline{E}mbeddings), and present two variants (Fig. \ref{fig:models} c-d):
\begin{description}
\item \textbf{\textit{Implicit} SAME (iSAME):} all the parameters $\theta$ are used for adaptation. This strategy makes use of the  \textit{implicit} feature-reuse factor of MAML, leading to parameters $\theta_{\text{GCN}}$ that are general across \textit{multi-task episodes}. 
\item \textbf{\textit{Explicit} SAME (eSAME):} only the head parameters $\theta_{\text{NC}}, \theta_{\text{GC}}, \theta_{\text{LP}}$ are used for adaptation (as done by ANIL). Contrary to iSAME, this strategy \textit{explicitly} aims at learning the parameters $\theta_{\text{GCN}}$ to be general across \textit{multi-task episodes} by only updating them in the outer loop.
\end{description}
The pseudocode in Algorithm \ref{meta_alg} is the same for both iSAME and eSAME. The difference between the methods is in which subset of the parameters $\theta$ is updated by the \texttt{ADAPT} function. In iSAME the \texttt{ADAPT} function will update the head and the backbone parameters ($\theta_{\text{GCN}}, \theta_{\text{NC}}, \theta_{\text{GC}}, \theta_{\text{LP}}$), while for eSAME only the head parameters ($\theta_{\text{NC}}, \theta_{\text{GC}}, \theta_{\text{LP}}$) will be updated.

\subsection{Connection between SAME and other Optimization-based Meta-Learning Methods}\label{samegeneralframework}

SAME is an instantiation of optimization-based meta-learning (in particular iSAME is an instantiation of MAML \citep{finn2017}, while eSAME of ANIL \citep{raghu2019rapid}) specially designed for learning multi-task representations. In particular SAME employs the following design choices:
\textbf{(1)} In SAME each episode is composed of multiple tasks (i.e. downstream applications). 
\textbf{(2)}  In SAME each task (both in the inner and in the outer loop) can involve only a subset of the parameters of the model. 
\textbf{(3)}  In SAME's inner loop, \textit{separate} adaptations are performed for each task in the episode. In the outer loop, the meta-objective defines how these multiple adaptations are combined for updating the initial representations of the parameters.

After training a model with SAME, an encoder is extracted and used to generate representations of the input that can then be fed to any machine learning model. As SAME is only used for training, no adaptation is performed at test time, and hence support and target sets are not required at test time.

\section{Experiments}
\label{sec:experiments}

Our goal is to assess the quality of the representations learned by models trained with SAME, and to study the impact of SAME's underlying components. In more detail, we aim to answer the following questions:
\begin{description}
\item[ \textbf{Q1:}] \textit{Do iSAME and eSAME lead to 
node embeddings that can be used to perform multiple downstream tasks with comparable (or better) performance than end-to-end single-task models?}
\item[ \textbf{Q2:}] \textit{Can node embeddings learned by a model trained with iSAME and eSAME be used for multiple tasks with comparable or better performance than classically trained (i.e., see Section~\ref{multihead}) multi-task models?}
\item[ \textbf{Q3:}] \textit{Do  iSAME and eSAME extract information that is not captured by the classical training procedure (i.e., see Section~\ref{multihead})?}
\item[ \textbf{Q4 \textit{(Ablation Study)}:}] \textit{What are the contributions of the different components of SAME's meta-learning procedure?}
\end{description}

Unless otherwise stated, accuracy (\%) is used for NC and GC, while ROC AUC (\%) is used for LP. (As a reminder, we use GC to refer to graph classification, NC for node classification, and LP for link prediction.)

\textbf{Datasets.} To perform multiple tasks, we consider datasets with graph labels, node attributes, and node labels from the TUDataset library \citep{Morris+2020}: ENZYMES \citep{PMID:14681450}, PROTEINS \citep{dobson}, DHFR and COX2 \citep{doi:10.1021/ci034143r}. ENZYMES is a dataset of protein structures belonging to six classes. PROTEINS is a dataset of chemical compounds with two classes (enzyme and non-enzyme). 
DHFR, and COX2 are datasets of chemical inhibitors which can be active or inactive. 

\textbf{Experimental Setup.} We perform a 10-fold cross validation, and average results across folds. 
To ensure a fair comparison, the same architecture is used for all training strategies. 
We set $\lambda^{(GC)} = \lambda^{(NC)} = \lambda^{(LP)} = 1$ as we noticed that weighting the losses did not provide significant benefits. Loss balancing techniques (e.g. \textit{uncertainty weights} \citep{8578879}, and \textit{gradnorm} \citep{chen2018gradnorm}) were tested, both with SAME and with the classical training procedure, but they did not result effective. This is  in accordance with recent works \cite{v2020revisiting,kurin2022defense} which observe that, when appropriate tuning is done, no method is significantly better than minimizing the sum of the task losses. For more information we refer to Appendix B, and we publicly release source code\footnote{\url{https://github.com/DavideBuffelli/SAME}}. 

\textbf{Q1:}
We train a model with SAME, on all multi-task combinations, and use the embeddings produced by the learned encoder as the input for a \textbf{linear classifier}. We compare against models with the same task-specific architecture trained in a classical supervised manner on a single task, and with a fine-tuning baseline. The latter is a model that has been trained on all three tasks, and then fine-tuned on two specific tasks. The idea is that the initial training on all tasks should lead the model towards the extraction of features that it would otherwise not consider (by only seeing 2 tasks). The fine-tuning process should then allow the model to use these features to target the specific tasks of interest. Results are shown in Table \ref{tableSAMEcombs}. 
The embeddings produced by the model learned with SAME in a multi-task setting achieve performance comparable to, and frequently even better than, end-to-end single-task models. In fact, the embeddings from SAME are never outperformed by more than 3\%, and in 50\% of the cases actually achieve higher performance. Moreover, the fine-tuning baseline 
is almost always outperformed by both single-task models, and our proposed methods. 
These results confirm that meta-learning is a powerful solution for multi-task representation learning on graphs.
\begin{table}
  \caption{Results for a single-task model trained in a classical supervised manner, a fine-tuned model (trained on all, and fine-tuned on two tasks), and a \textbf{linear} classifier trained on node embeddings generated by a model trained with our strategies (iSAME, eSAME) in a multi-task setting.}\label{tableSAMEcombs}
  \centering
  \resizebox{\linewidth}{!}{%
  \begin{tabular}{ccccccc}
    \hline
    \multicolumn{3}{c}{\textbf{Task}} & \multicolumn{4}{c}{\textbf{Dataset}} \\
                GC & NC & LP                 & ENZYMES & PROTEINS & DHFR & COX2 \\
               &&& GC \phantom{ /} NC \phantom{/ } LP & GC \phantom{ /} NC \phantom{/ } LP & GC \phantom{ /} NC \phantom{/ } LP & GC \phantom{ /} NC \phantom{/ } LP\\
    \hline
    \multicolumn{7}{c}{\textbf{Classical End-to-End Training}}\\
    \hline
     \checkmark &  &                    & 51.6 \phantom{/} \phantom{00.0} \phantom{/} \phantom{00.0} & 73.3 \phantom{/} \phantom{00.0} \phantom{/} \phantom{00.0} & 71.5 \phantom{/} \phantom{00.0} \phantom{/} \phantom{00.0} & 76.7 \phantom{/} \phantom{00.0} \phantom{/} \phantom{00.0}\\
                        & \checkmark &                                & \phantom{00.0} \phantom{/} 87.5 \phantom{/} \phantom{00.0} & \phantom{00.0} \phantom{/} 72.3 \phantom{/} \phantom{00.0} & \phantom{00.0} \phantom{/} 97.3 \phantom{/} \phantom{00.0} & \phantom{00.0} \phantom{/} 96.4 \phantom{/} \phantom{00.0}\\
                        &  & \checkmark                                & \phantom{00.0} \phantom{/} \phantom{00.0} \phantom{/} 75.5 & \phantom{00.0} \phantom{/} \phantom{00.0} \phantom{/} 85.6 & \phantom{00.0} \phantom{/} \phantom{00.0} \phantom{/} 98.8 & \phantom{00.0} \phantom{/} \phantom{00.0} \phantom{/} 98.3\\
    \hline
    \multicolumn{7}{c}{\textbf{Fine-Tuning}}\\
    \hline
    \checkmark & \checkmark &                     & 48.3 \phantom{/} 85.3 \phantom{/} \phantom{00.0} & 73.6 \phantom{/} 72.0 \phantom{/} \phantom{00.0} & 66.4 \phantom{/} 92.4 \phantom{/} \phantom{00.0} & 80.0 \phantom{/} 92.3 \phantom{/} \phantom{00.0}\\
     \checkmark &  & \checkmark                                                       & 49.3 \phantom{/} \phantom{00.0} \phantom{/} 71.6 & 69.6 \phantom{/} \phantom{00.0} \phantom{/} 80.7 & 65.3 \phantom{/} \phantom{00.0} \phantom{/} 58.9 & 80.2 \phantom{/} \phantom{00.0} \phantom{/} 50.9\\
     & \checkmark  & \checkmark                                                       & \phantom{00.0} \phantom{/} 87.7 \phantom{/} 73.9 & \phantom{00.0} \phantom{/} 80.4 \phantom{/} 81.5 & \phantom{00.0} \phantom{/} 80.7 \phantom{/} 56.6 & \phantom{00.0} \phantom{/} 87.4 \phantom{/} 52.3\\
    \hline
    \multicolumn{7}{c}{\textbf{iSAME (ours)}}\\
    \hline
     \checkmark & \checkmark &                     & 50.1 \phantom{/} 86.1 \phantom{/} \phantom{00.0} & 73.1 \phantom{/} 76.6 \phantom{/} \phantom{00.0} & 71.6 \phantom{/} 94.8 \phantom{/} \phantom{00.0} & 75.2 \phantom{/} 95.4 \phantom{/} \phantom{00.0}\\
     \checkmark &  & \checkmark                                                       & 50.7 \phantom{/} \phantom{00.0} \phantom{/} 83.1 & 73.4 \phantom{/} \phantom{00.0} \phantom{/} 85.2 & 71.6 \phantom{/} \phantom{00.0} \phantom{/} 99.2 & 77.5 \phantom{/} \phantom{00.0} \phantom{/} 98.9\\
     & \checkmark  & \checkmark                                                       & \phantom{00.0} \phantom{/} 86.3 \phantom{/} 83.4 & \phantom{00.0} \phantom{/} 79.4 \phantom{/} 87.7 & \phantom{00.0} \phantom{/} 96.5 \phantom{/} 99.3 & \phantom{00.0} \phantom{/} 95.5 \phantom{/} 99.0\\
     \checkmark & \checkmark  & \checkmark                                    & 50.0 \phantom{/} 86.5 \phantom{/} 82.3 & 71.4 \phantom{/} 76.6 \phantom{/} 87.3 & 71.2 \phantom{/} 95.5 \phantom{/} 99.5 & 75.4 \phantom{/} 95.2 \phantom{/} 99.2\\
    \hline
    \multicolumn{7}{c}{\textbf{eSAME (ours)}}\\
    \hline
     \checkmark & \checkmark &                     & 51.7 \phantom{/} 86.1 \phantom{/} \phantom{00.0} & 71.5 \phantom{/} 79.2 \phantom{/} \phantom{00.0} & 70.1 \phantom{/} 95.7 \phantom{/} \phantom{00.0} & 75.6 \phantom{/} 95.5 \phantom{/} \phantom{00.0}\\
     \checkmark &  & \checkmark                                                       & 51.9 \phantom{/} \phantom{00.0} \phantom{/} 80.1 & 71.7 \phantom{/} \phantom{00.0} \phantom{/} 85.4 & 70.1 \phantom{/} \phantom{00.0} \phantom{/} 99.1 & 77.5 \phantom{/} \phantom{00.0} \phantom{/} 98.8\\
     & \checkmark  & \checkmark                                                       & \phantom{00.0} \phantom{/} 86.7 \phantom{/} 82.2 & \phantom{00.0} \phantom{/} 80.7 \phantom{/} 86.3 & \phantom{00.0} \phantom{/} 96.6 \phantom{/} 99.4 & \phantom{00.0} \phantom{/} 95.6 \phantom{/} 99.1\\
     \checkmark & \checkmark  & \checkmark                                    & 51.5 \phantom{/} 86.3 \phantom{/} 81.1 & 71.3 \phantom{/} 79.6 \phantom{/} 86.8 & 70.2 \phantom{/} 95.3 \phantom{/} 99.5 & 77.7 \phantom{/} 95.7 \phantom{/} 98.8\\
    \hline
  \end{tabular}
  }
\end{table}

\begin{table}[t]
  \caption{$\Delta_m$ (\%) results for a classically trained multi-task model (Cl), a fine-tuned model (FT; trained on all three tasks and fine-tuned on two) and a \textbf{linear} classifier trained on the node embeddings generated by a model trained with our meta-learning strategies (iSAME, eSAME) in a multi-task setting.}\label{tableQ3}
  \centering
  \resizebox{\linewidth}{!}{%
  \begin{tabular}{cccccccc}
    \hline
    \multicolumn{3}{c}{\textbf{Task}} & \textbf{Model} & \multicolumn{4}{c}{\textbf{Dataset}} \\
               GC & NC & LP     &            & ENZYMES & PROTEINS & DHFR & COX2 \\
    \hline
    \multirow{4}{*}{\checkmark} & \multirow{4}{*}{\checkmark} & \multirow{4}{*}{ } & Cl & $-0.1 \pm 0.5$ & $4.0 \pm 1.0$ & $-0.3 \pm 0.2$ & $0.5 \pm 0.1$\\
                                                     &&                                                                    & FT & $-4.5 \pm 1.2$ & $0.1 \pm 0.5$ & $-7.4 \pm 1.4$ & $0.1 \pm 0.4$\\
                                                     &&                                                                    & iSAME & $-2.3 \pm 0.9$ & $2.7 \pm 1.5$ & $-1.2 \pm 0.4$ & $-1.6 \pm 0.2$\\
                                                     &&                                                                    & eSAME & $-0.8 \pm 0.8$ & $3.2 \pm 1.4$ & $-1.8 \pm 0.3$ & $-1.2 \pm 0.3$\\
    \hline
    \multirow{4}{*}{\checkmark} & \multirow{4}{*}{ } & \multirow{4}{*}{\checkmark} & Cl & $-25.3 \pm 3.2$ & $-5.3 \pm 1.2$ & $-28.3 \pm 4.3$ & $-21.4 \pm 3.4$\\
                                                         &&                                                                    & FT & $-5.1 \pm 1.9$ & $-5.4 \pm 1.5$ & $-24.5 \pm 3.7$ & $-22.6 \pm 3.8$\\
                                                     &&                                                                    & iSAME & $4.1 \pm 0.5$ & $-0.2 \pm 0.9$ & $0.2 \pm 3.2$ & $0.2 \pm 0.5$\\
                                                     &&                                                                    & eSAME & $3.2 \pm 0.4$ & $-1.2 \pm 1.1$ & $-0.7 \pm 3.4$ & $-0.8 \pm 0.7$\\
    \hline
    \multirow{4}{*}{ } & \multirow{4}{*}{\checkmark} & \multirow{4}{*}{\checkmark} & Cl & $7.2 \pm 2.7$ & $6.8 \pm 0.9$ & $-29.1 \pm 7.7$ & $-28.2 \pm 4.5$\\
                                                         &&                                                                    & FT & $-1.0 \pm 0.3$ & $3.1 \pm 1.2$ & $-28.9 \pm 6.4$ & $-28.3 \pm 4.2$\\
                                                     &&                                                                    & iSAME & $4.4 \pm 1.1$ & $6.1 \pm 1.0$ & $-0.1 \pm 6.2$ & $-0.6 \pm 2.5$\\
                                                     &&                                                                    & eSAME & $3.9 \pm 1.3$ & $6.1 \pm 1.1$ & $0.1 \pm 6.4$ & $-0.6 \pm 2.6$\\
    \hline
    \multirow{3}{*}{\checkmark} & \multirow{3}{*}{\checkmark} & \multirow{3}{*}{\checkmark} & Cl & $ 1.6 \pm 1.3$ & $2.9 \pm 0.3$ & $-18.9 \pm 2.3$ & $-16.9 \pm 3.1$\\
                                                     &&                                                                    & iSAME & $1.5 \pm 1.0$ & $2.2 \pm 0.2$ & $-0.5 \pm 1.4$ & $-0.9 \pm 1.3$\\
                                                     &&                                                                    & eSAME & $1.8 \pm 0.9$ & $2.8 \pm 0.2$ & $-1.0 \pm 1.7$ & $-0.4 \pm 1.2$\\
    \hline
  \end{tabular}
  }
\end{table}
\textbf{Q2:} 
We train the same multi-task model, both in the classical supervised manner (see Section \ref{multihead}), and with our proposed approaches, on all multi-task combinations. For our approaches, a \textbf{linear classifier} is then trained on top of the node embeddings produced by the learned encoder. We further consider the fine-tuning baseline introduced in \textbf{Q1}. The multi-task performance ($\Delta_m$) metric \citep{8954118} is used, defined as the average per-task drop with respect to the single-task baseline:
$\Delta_m = \frac{1}{T} \sum_{i=1}^{T} \left(M_{m,i} - M_{b,i} \right) / M_{b,i},$
where $M_{m,i}$ is the value of a task's metric for the multi-task model, and $M_{b,i}$ is the value for the baseline. Results are shown in Table \ref{tableQ3}. Multi-task models usually achieve lower performance than specialized single-task ones. Moreover, \textbf{linear} classifiers trained on the embeddings generated by a model trained with SAME are not only comparable, but in many cases significantly superior to classically trained multi-task models. In fact, a multi-task model trained in a classical manner is highly sensible to the tasks that are being learned (e.g. GC and LP negatively interfere with each other in every dataset), while our methods are much less sensible. For instance, the former has a worst-case average drop in performance of 29\%, while our method has a worst-case average drop of less than 3\%. Finally, the fine-tuning baseline generally performs worse than classically trained models, confirming that transferring knowledge in multi-task settings is not easy. 

\begin{figure}[h]
\centering
 \subfloat{\includegraphics[width=0.75\linewidth]{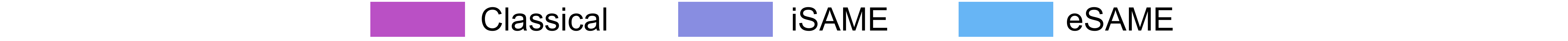}}\\
 \vspace{-1mm}
  \subfloat[      (a)]{\includegraphics[width=0.2\linewidth]{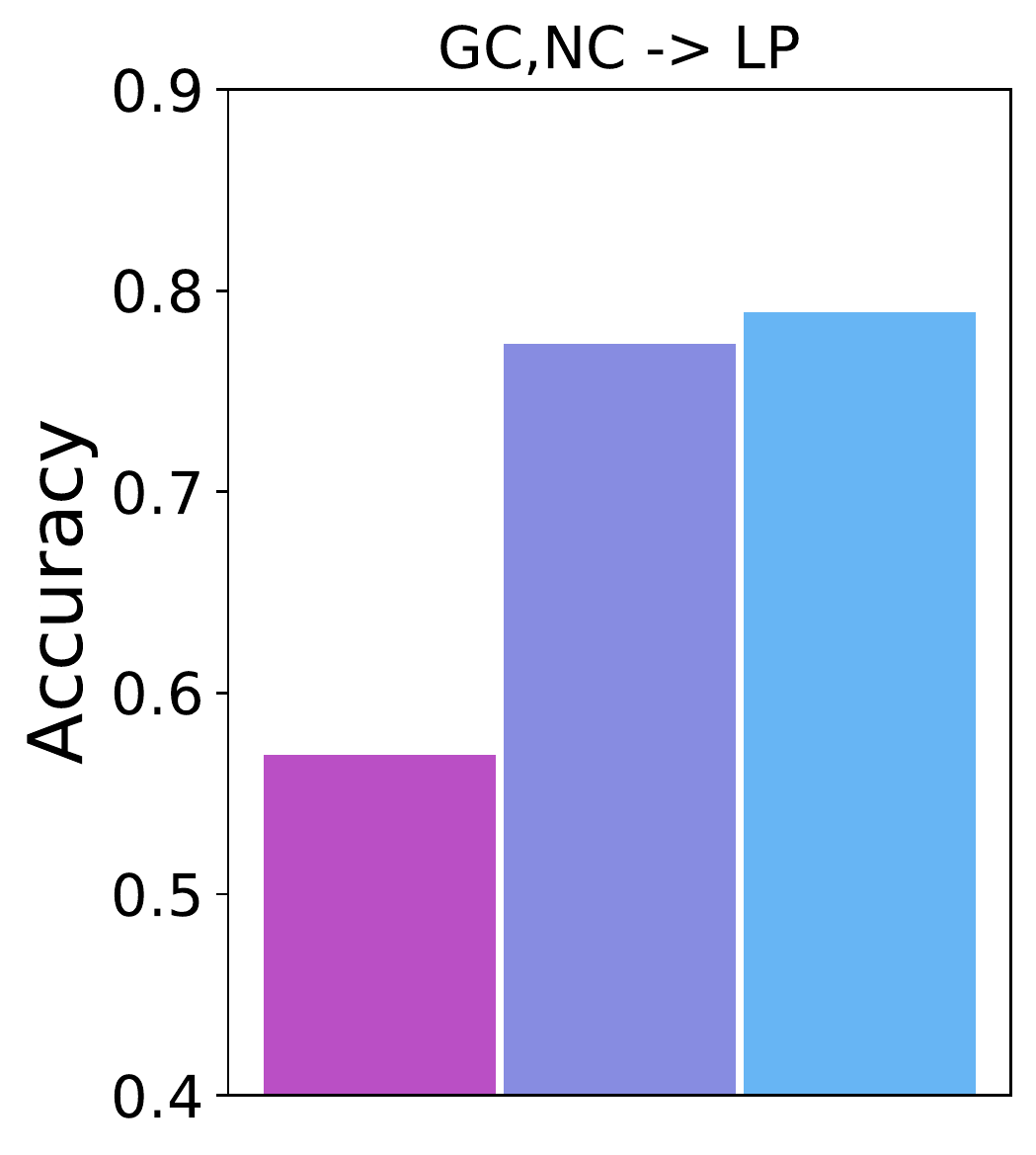}}
  \subfloat[      (b)]{\includegraphics[width=0.2\linewidth]{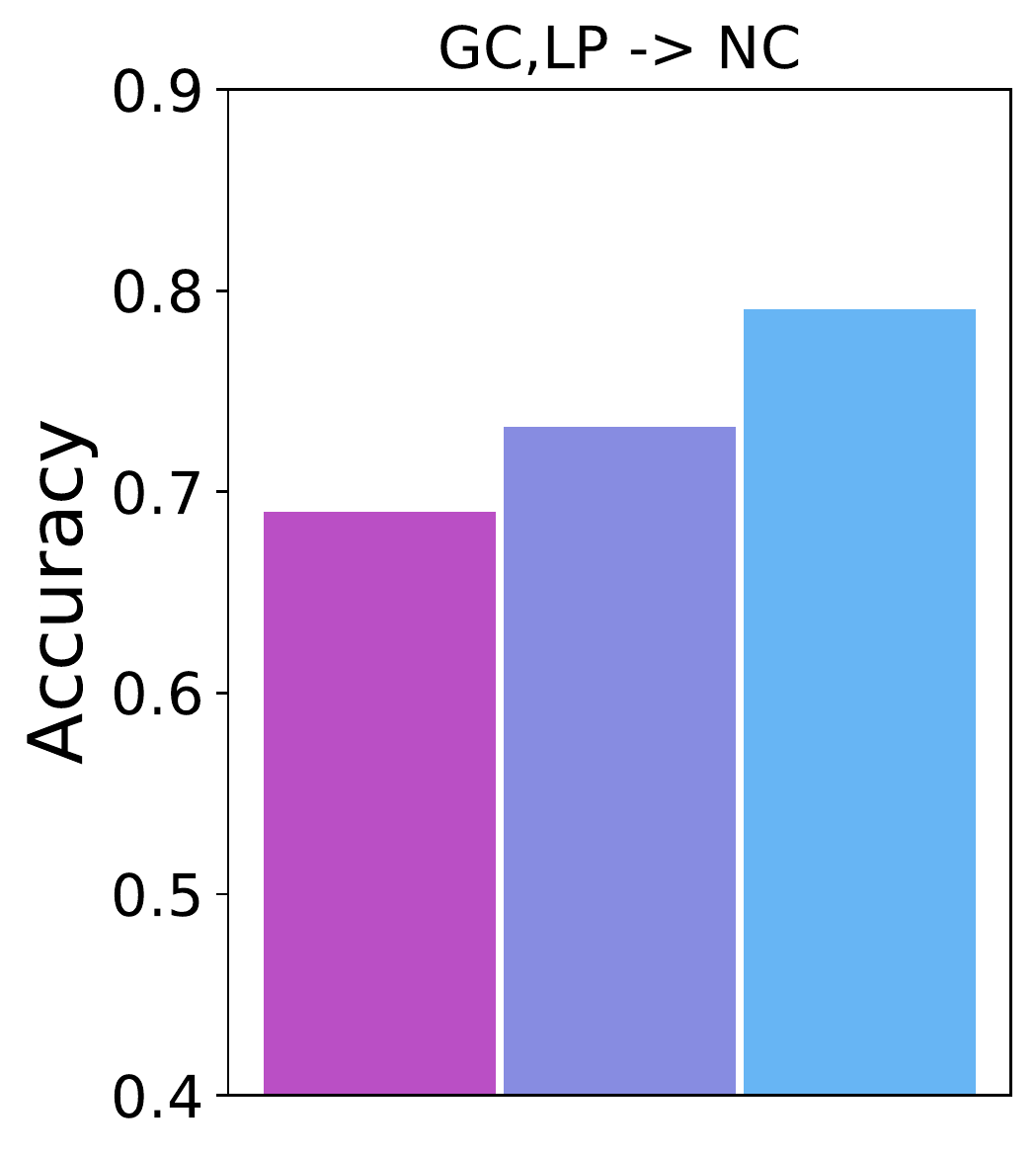}}
  \subfloat[      (c)]{\includegraphics[width=0.2\linewidth]{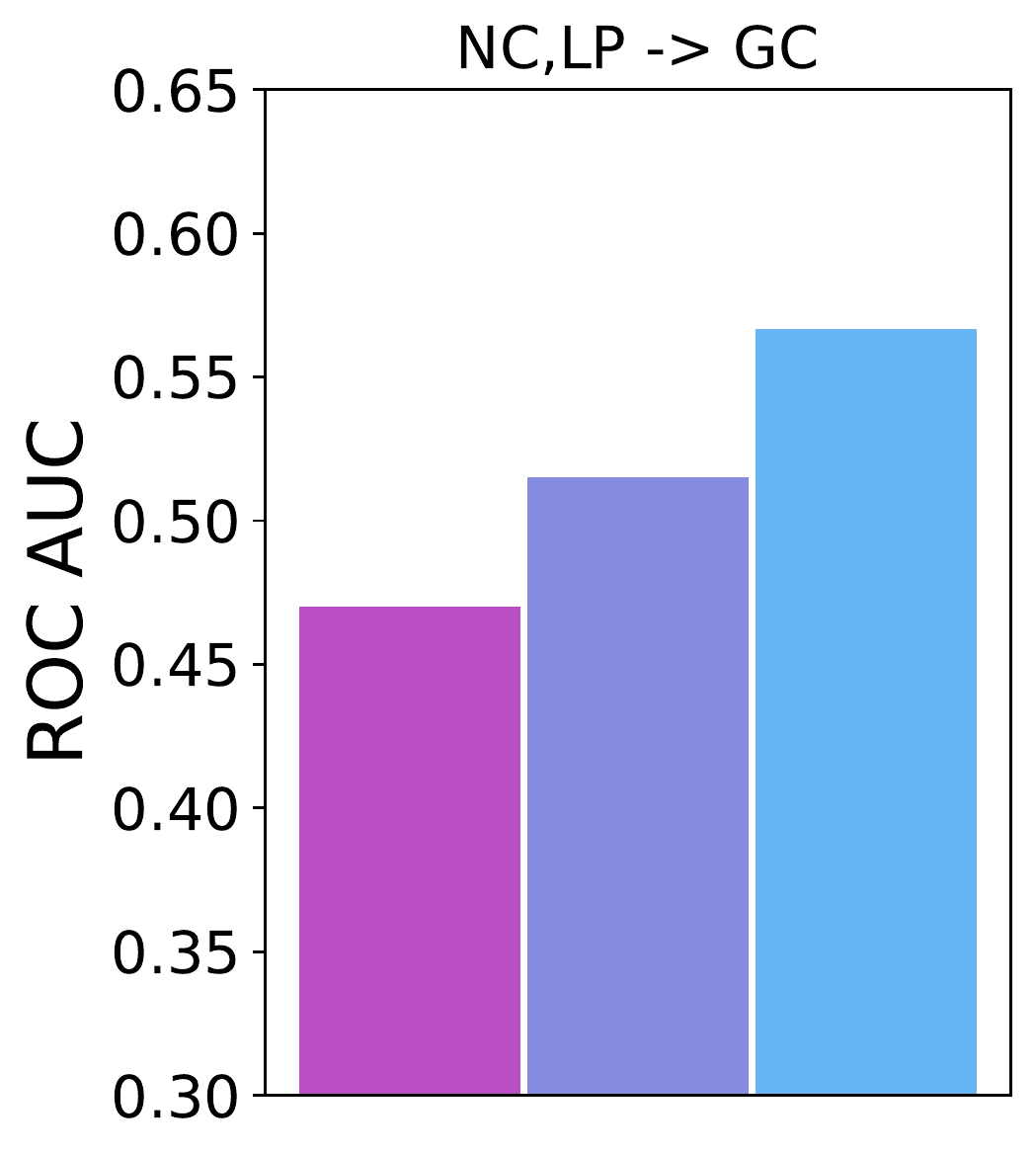}}
  \caption{Results for model, trained on the embeddings generated by a multi-task model, performing a task that was not seen during training. ``$x,y$-\textgreater$z$'' indicates that $x,y$ are the tasks used for training the multi-task model, and $z$ is the new task.}
  \label{fig:transfer_multitask_embeddings}
\end{figure}

\textbf{Q3:}
We train a multi-task model, and then train a new simple network (with the same architecture as the heads described in Section \ref{architecture}), which is refer to as \textit{classifier}, on the embeddings generated by the multi-task model to perform a task that was not seen during training. We compare the performance of the classifier on the embeddings generated by a model trained in a classical manner, and with SAME. Intuitively, this tests gives us a way to analyse if the embeddings generated by a model trained with SAME contain ``more information'' than embeddings generated by a model trained in a classical manner. Results on the ENZYMES dataset are shown in Fig. \ref{fig:transfer_multitask_embeddings}. Interestingly, the embeddings generated by a model trained with SAME lead to at least 10\% higher performance. We observe an analogous trend on the other datasets (full results are in Appendix C).

\begin{figure*}[h]
\centering
 \subfloat{\includegraphics[width=0.70\linewidth]{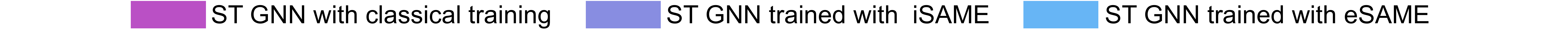}}\\
 \vspace{-1mm}
  \subfloat[      (a)]{\includegraphics[width=0.29\textwidth]{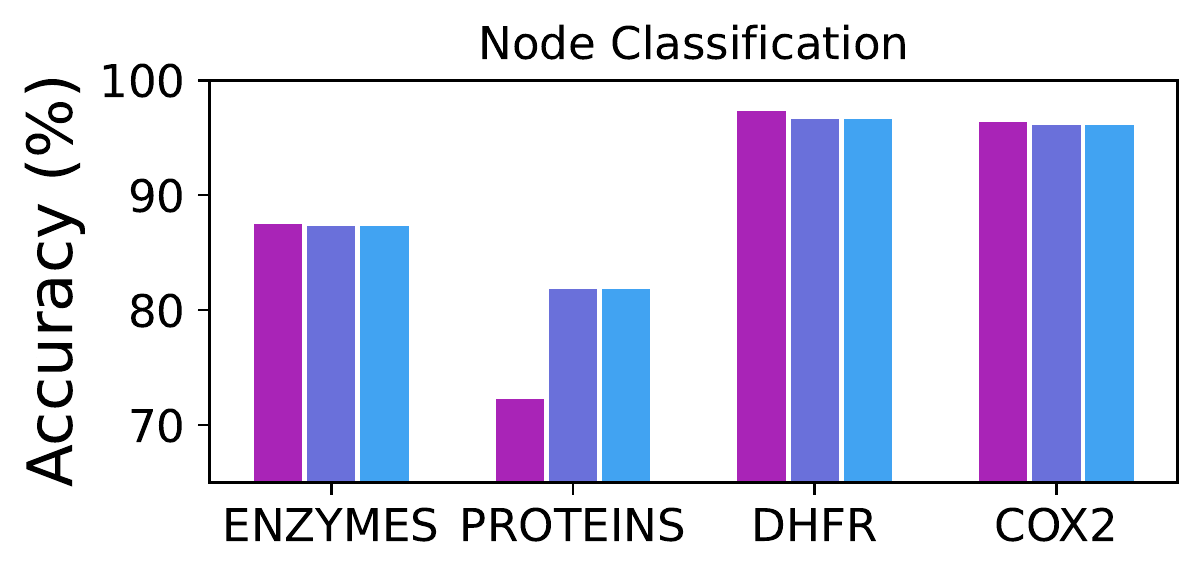}}
  \subfloat[      (b)]{\includegraphics[width=0.29\textwidth]{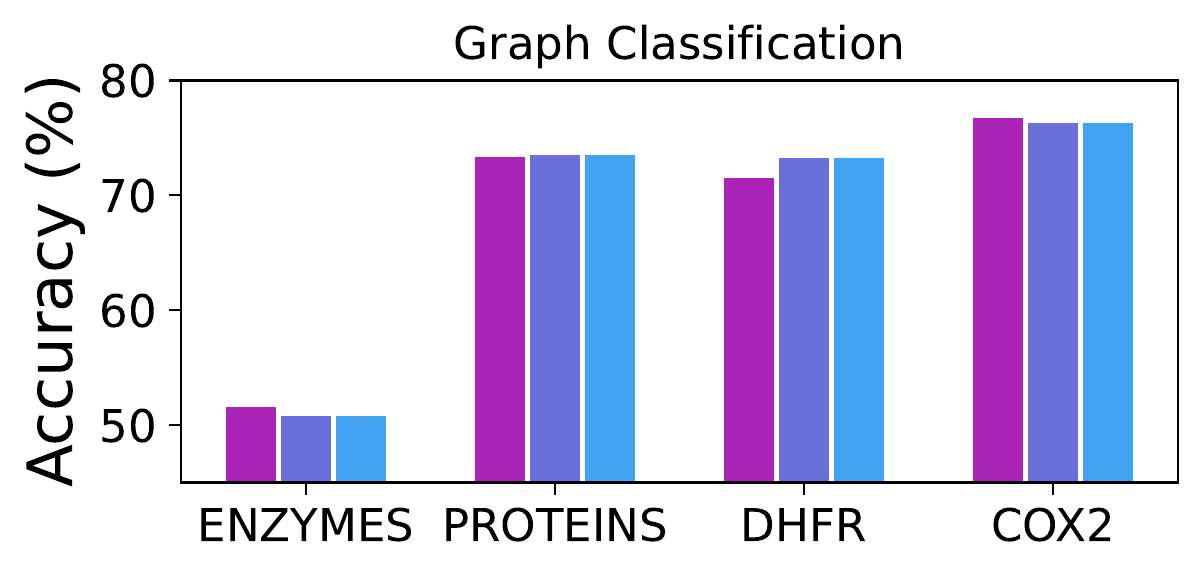}}
  \subfloat[      (c)]{\includegraphics[width=0.29\textwidth]{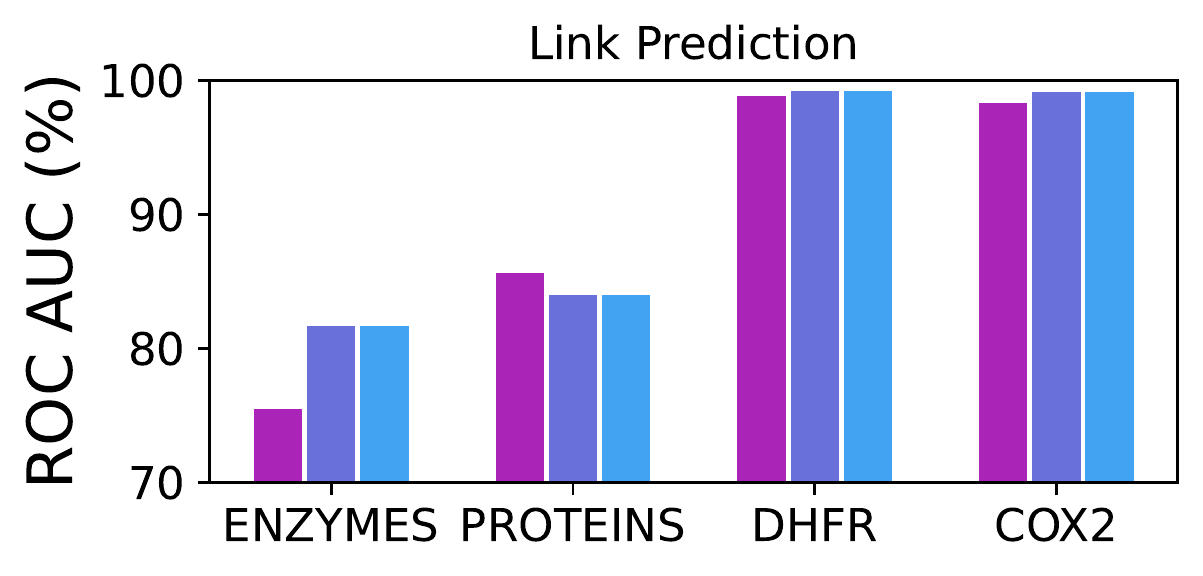}}
  \caption{Results for a Single-Task GNN model (ST GNN) trained with the classical procedure, and a \textbf{linear} classifier trained on the embeddings generated by a model trained with an ablated ``single-task'' version of SAME.}
  \label{fig:singeltaskablation}
\end{figure*}

\textbf{Q4 \textit{(Ablation Study)}:}
SAME's meta-learning procedure has two main ingredients: 
\begin{description}
\item \textbf{\textit{(1)}} the design of support and target sets, 
to encourage generalization by 
mimicking
training and validation sets (see Section \ref{epdesign}).
\item \textbf{\textit{(2)}} the separate multiple single-task adaptations performed in the inner loop, which relieve the model from having to learn to solve all the tasks \textit{concurrently} (Section \ref{adaptations}). 
\end{description}
To better understand the importance and contribution of each component we perform two experiments, for which results are presented below (the full results can be found in Appendix E). 

First, we isolate the contribution of \textbf{\textit{(1)}} by applying iSAME and eSAME in a \textit{single-task} setting (i.e., the same \emph{single} task is performed in both inner and outer loops), with episodes following the generalization-encouraging design proposed in Section \ref{epdesign}. Notice that this is like applying the original MAML and ANIL training procedures with our design of support and target sets. 
In this experiment, for every task, we train a \textbf{linear classifier} on top of the embeddings produced by a model trained with ``single-task'' iSAME and eSAME, and compare against a network with the same architecture trained in a classical end-to-end manner. Results are shown in Fig. \ref{fig:singeltaskablation}. For all three tasks, a \textbf{linear} classifier on the embeddings produced by a model trained with our methods achieves comparable, if not superior, performance to an end-to-end model. In fact, the linear classifier is never outperformed by more than 2\%, and it can outperform the classical end-to-end model by up to 12\%. We believe this unexpected outcome is particularly interesting, and hints that episodic training procedures can be used to learn better representations.

Second, we investigate the benefits of \textbf{\textit{(2)}} by removing the separate multiple single-task adaptations of SAME and performing \emph{all} tasks (i.e., GC, NC, and LP) concurrently both in the inner and outer loop. This leads to a simple \emph{concurrent multi-task version} of the conventional training procedure of MAML and ANIL, but with our support and target set design. 
For this experiment, we evaluate the ablated versions of SAME on the same procedure of \textbf{Q2} and \textbf{Q3}, and compare against the results of iSAME and eSAME. The results from the ablated version are not significantly different from those of \emph{non-ablated} iSAME and eSAME (see Appendix D).

From these experiments we draw two conclusions. \textbf{\textit{(i)}} The \textit{generalization-encouraging} design of support and target sets is what allows SAME to reach performance on multiple tasks that are comparable to specialised single-task models trained in a classical manner. \textbf{\textit{(ii)}} The separate \textit{multiple single-task adaptations} that are performed in the inner loop of iSAME and eSAME allow the models to reach the same performance of a version of SAME where all tasks are performed concurrently on all graphs, hence increasing the learning efficiency by not requiring labels for each task on every graph.

\section{Related Work}\label{rel_work}
GNNs, MTL, and meta-learning are very active areas of research. We highlight works that are at the intersection of these subjects, and point the interested reader to comprehensive reviews of each field. To the best of our knowledge there is no work using meta-learning to train a model for graph MTL, or proposing a GNN performing graph classification, node classification, and link prediction \textit{concurrently}.

\textbf{Graph Neural Networks.}
GNNs have a long history \citep{4700287}, but in the past few years the field has grown exponentially \citep{Chami2020MachineLO,Wu_2020}.
The first popular GNN approaches were based on filters in the graph spectral domain \citep{7974879}, and presented many challenges including high computational complexity. \citet{cnn_graph} introduced ChebNet, which uses Chebyshev polynomials to produce localized and efficient filters in the graph spectral domain. Graph Convolutional Networks \citep{kipf2017semi} introduced a localized first-order approximation of spectral graph convolutions which was then extended to include attention mechanisms  \citep{velickovic2018graph}. Recently, \citet{xu2018how} provided theoretical ground for the expressivity of GNNs. 

\textbf{Multi-Task Learning.}
Works at the intersection of MTL and GNNs have mostly focused on multi-head architectures. These models are composed of a series of GNN layers followed by multiple heads (i.e. independent neural network layers) that perform the desired downstream tasks. In this category, \citet{Montanari_2019} propose a model for the prediction of physico-chemical properties. \citet{holtz4multi} and \citet{Xie_2020} propose multi-task models for concurrently performing node and graph classification. Finally, \citet{Avelar_2019} introduce a multi-head GNN for learning multiple graph centrality measures, and \citet{li-ji-2019-syntax} propose a MTL method for the extraction of multiple biomedical relations. Other related work includes \citep{haonan2019graph} which introduces a model that can be trained for several tasks singularly, hence, unlike the previously mentioned approaches and our proposed method, it can not perform multiple tasks concurrently.
There are also some works that use GNNs as a tool for MTL: \citet{Liu2019LearningMC} use GNNs to allow communication between tasks, while \citet{10.5555/3327345.3327479} use GNNs to estimate the test error of a MTL model. 
In summary, the current literature on graph MTL has focused on multi-head architectures that are trained end-to-end. 
In this work we tackle the graph representation learning scenario in which the node embeddings are used for multiple tasks, and propose the use of meta-learning for training a GNN for this setting.
We further mention the work by \citet{10.3389/fnins.2019.01387} that considers the task of generating ``general'' node embeddings, however their method is not based on GNNs, does not consider node attributes (unlike our method), and is not focused on the three most common graph related tasks.
For an exhaustive review of deep MTL techniques we refer the reader to \citet{v2020revisiting}.

\textbf{Meta-Learning.}
Meta-Learning consists in \textit{learning to learn}. Many methods have been proposed (see the review by \citet{hospedales2020metalearning}), specially in the area of \textit{few-shot learning}. 
 \citet{garcia2017few} frame the few-shot learning problem with a partially observed graphical model and use GNNs as an inference algorithm. 
\citet{liu2019GPN} use GNNs to propagate messages between class prototypes and improve existing few-shot learning methods, while \citet{10.1145/3394486.3403230} use GNNs to introduce domain-knowledge in the form of graphs.
There are also several works that use meta-learning to train GNNs in few-shot learning scenarios with applications to node classification \citep{10.1145/3357384.3358106,yeaofewshot2020}, edge labelling \citep{Kim2019EdgeLabelingGN}, link prediction \citep{Alet2019NeuralRI,bose2019meta}, and graph regression \citep{nguyen2020}. Finally, other combinations of meta-learning and GNNs involve adversarial attacks \citep{zugner_adversarial_2019} and active learning \citep{madhawa2020}. 


\section{Conclusions}\label{conclusions}
This work introduces the use of meta-learning as a training strategy for graph representation learning in multi-task settings.
We find that our method leads to models that produce ``more general'' node embeddings.
In fact, our results show that the embeddings produced by a model trained with our technique can be used to perform graph classification, node classification, and link prediction, with comparable or better performance than separate single-task end-to-end supervised models.
Furthermore, we find that the embeddings generated by a model trained with our procedure lead to higher performance on downstream tasks that were not seen during training, and that the episodic training procedure leads to better embeddings even in the single-task setting.
We believe this work can be of interest to the community as it explores the under-studied area of multi-task representation learning on graphs, and further introduces a method built on optimization-based meta-learning (inheriting the properties of being \textit{model-agnostic}), which can be adapted to other domains as future work.
Another interesting direction is to incorporate more advanced meta-learning strategies like \cite{rajeswaran2019meta}.

{\small 
\section*{Acknowledgment}
This work is supported, in part, by the Italian Ministry of Education, University and Research (MIUR), under PRIN Project n. 20174LF3T8 ``AHeAD'' and the initiative ``Departments of Excellence'' (Law 232/2016), and by University of Padova under project ``SID 2020: RATED-X''. 

\setlength{\bibsep}{0pt plus 0.3ex}
\bibliographystyle{abbrvnat}
\bibliography{references}
}

\appendix
\section{Episode Design Algorithm}\label{episodealgo}
Algorithm \ref{epi_alg} contains the procedure for the creation of the episodes for our meta-learning procedures. The algorithm takes as input a batch of graphs (with graph labels, node labels, and node features) and the loss function balancing weights, and outputs a \textit{multi-task episode}. We assume that each graph has a set of attributes that can be accessed with a \textit{dot-notation} (like in most object-oriented programming languages).

Notice how the episodes are created so that \textbf{only one task is performed on each graph} (which implies that we only need labels for one task for each graph). This is important as in the inner loop of our meta-learning procedure, the learner adapts and tests the adaptated parameters on one task at a time. The outer loop then updates the parameters, optimizing for a representation that leads to fast \textit{single-task adaptation}. This procedure bypasses the problem of learning parameters that \textit{directly} solve multiple tasks, which can be very challenging. 

Another important aspect to notice is that the support and target sets are designed as if they were the training and validation splits for training a single-task model with the classical procedure. This way the meta-objective becomes to train a model that can generalize well. 
\begin{algorithm*}[h!]
\caption{Episode Design Algorithm}\label{epi_alg}
\begin{algorithmic}
\STATE {\bfseries Input:} Batch of $n$ randomly sampled graphs $\mathcal{B} = \{ \mathcal{G}_1, .., \mathcal{G}_n \}$ \\ Loss weights $\lambda^{(GC)}, \lambda^{(NC)}, \lambda^{(LP)} \in [0,1]$
\STATE {\bfseries Output:} Episode $\mathcal{E}_i = ( \mathcal{L}^{(m)}_{\mathcal{E}_{i}},  \mathcal{S}^{(m)}_{\mathcal{E}_{i}},   \mathcal{T}^{(m)}_{\mathcal{E}_{i}} )$ \;
  \STATE$\mathcal{B}^{(GC)}, \mathcal{B}^{(NC)}, \mathcal{B}^{(LP)} \leftarrow$ equally divide the graphs in $\mathcal{B}$ in three sets \;
  \STATE
  \STATE
 \COMMENT{Graph Classification}
  \STATE$\mathcal{S}^{(\text{GC})}_{\mathcal{E}_{i}}, \mathcal{T}^{(\text{GC})}_{\mathcal{E}_{i}} \leftarrow$ randomly divide $\mathcal{B}^{(GC)}$ with a 60/40 split\;
   \STATE
   \STATE
  \COMMENT{Node Classification}
  \FOR{$\mathcal{G}_i$  {\bfseries in} $\mathcal{B}^{(NC)}$}
  \STATE $\text{\texttt{num\_labelled\_nodes}} \leftarrow \mathcal{G}_{i}\text{\texttt{.num\_nodes}} \times  0.3$ \;
  \STATE $\mathcal{N} \leftarrow$ divide nodes per class, then iteratively randomly sample one node per class without replacement and add it to $\mathcal{N}$ until $\lvert \mathcal{N} \vert = \text{\texttt{num\_labelled\_nodes}}$ \;
  \STATE $\mathcal{G}_{i}^{\prime} \leftarrow \text{\texttt{copy}}(\mathcal{G}_i)$ \;
  \STATE $\mathcal{G}_{i}\text{\texttt{.labelled\_nodes}} \leftarrow \mathcal{N}$; $\quad \mathcal{G}_{i}^{\prime}\text{\texttt{.labelled\_nodes}} \leftarrow \mathcal{G}_{i}\text{\texttt{.nodes}} \setminus \mathcal{N}$ \;
  \STATE $\mathcal{S}^{(NC)}_{\mathcal{E}_{i}}\text{\texttt{.add}}(\mathcal{G}_{i})$; $\quad \mathcal{T}^{(NC)}_{\mathcal{E}_{i}}\text{\texttt{.add}}(\mathcal{G}_{i}^{\prime})$
   \ENDFOR
 
 \STATE
   \STATE
 \COMMENT{Link Prediction}
  \FOR{$\mathcal{G}_i$  {\bfseries in} $\mathcal{B}^{(LP)}$}
  \STATE $E_{i}^{(N)} \leftarrow$ randomly pick negative samples (edges that are not in the graph; possibly in the same number as the number of edges in the graph) \;
  \STATE $E_{i}^{1,(N)}, E_{i}^{2,(N)} \leftarrow$ divide $E_{i}^{(N)}$ with an $80/20$ split \;
  \STATE $E_{i}^{(P)} \leftarrow$ randomly remove $20\%$ of the edges in $\mathcal{G}_{i}$ \;
  \STATE $\mathcal{G}_{i}^{\prime (1)} \leftarrow \mathcal{G}_{i}$ removed of $E_{i}^{(P)}$ \;
  \STATE $\mathcal{G}_{i}^{\prime (2)} \leftarrow \text{\texttt{copy}}( \mathcal{G}_{i}^{\prime (1)})$ \;
  \STATE $\mathcal{G}_{i}^{\prime (1)} \text{\texttt{.positive\_edges}} \leftarrow \mathcal{G}_{i}^{\prime (1)} \text{\texttt{.edges}}$; $\quad \mathcal{G}_{i}^{\prime (2)} \text{\texttt{.positive\_edges}} \leftarrow E_{i}^{(P)} $ \;
  \STATE $\mathcal{G}_{i}^{\prime (1)} \text{\texttt{.negative\_edges}} \leftarrow E_{i}^{1,(N)}$; $\quad \mathcal{G}_{i}^{\prime (2)} \text{\texttt{.negative\_edges}} \leftarrow E_{i}^{2,(N)}$ \;
  \STATE $\mathcal{S}^{(LP)}_{\mathcal{E}_{i}}\text{\texttt{.add}}(\mathcal{G}_{i}^{\prime (1)})$; $\quad \mathcal{T}^{(LP)}_{\mathcal{E}_{i}}\text{\texttt{.add}}(\mathcal{G}_{i}^{\prime (2)})$
  \ENDFOR
 
 \STATE
   \STATE
 \STATE $\mathcal{S}^{(m)}_{\mathcal{E}_{i}} \leftarrow \{ \mathcal{S}^{(\text{GC})}_{\mathcal{E}_{i}}, \mathcal{S}^{(\text{NC})}_{\mathcal{E}_{i}}, \mathcal{S}^{(\text{LP})}_{\mathcal{E}_{i}} \}$ \;
  \STATE $\mathcal{T}^{(m)}_{\mathcal{E}_{i}} \leftarrow \{ \mathcal{T}^{(\text{GC})}_{\mathcal{E}_{i}}, \mathcal{T}^{(\text{NC})}_{\mathcal{E}_{i}}, \mathcal{T}^{(\text{LP})}_{\mathcal{E}_{i}} \}$ \;
 \STATE $\mathcal{L}^{(\text{GC})}_{\mathcal{T}_{i}} \leftarrow $ \texttt{Cross-Entropy}$(\cdot)$; $\quad \mathcal{L}^{(\text{NC})}_{\mathcal{T}_{i}} \leftarrow $ \texttt{Cross-Entropy}$(\cdot)$ \;
 \STATE $\mathcal{L}^{(\text{LP})}_{\mathcal{T}_{i}} \leftarrow $ \texttt{Binary Cross-Entropy}$(\cdot)$ \;
 \STATE $ \mathcal{L}^{(m)}_{\mathcal{E}_{i}} = \lambda^{(GC)} \mathcal{L}^{(\text{GC})}_{\mathcal{T}_{i}} + \lambda^{(NC)} \mathcal{L}^{(\text{NC})}_{\mathcal{T}_{i}} + \lambda^{(LP)} \mathcal{L}^{(\text{LP})}_{\mathcal{T}_{i}}$ \;
 \STATE \textbf{Return} $\mathcal{E} = ( \mathcal{L}^{(m)}_{\mathcal{E}_{i}},  \mathcal{S}^{(m)}_{\mathcal{E}_{i}},   \mathcal{T}^{(m)}_{\mathcal{E}_{i}} )$
 \end{algorithmic}
\end{algorithm*}

\section{Additional Experimental Details}\label{expdetails}
In this section we provide additional information on the implementation of the models used in our experimental section. We implement our models using PyTorch \citep{NEURIPS2019_9015}, PyTorch Geometric \citep{Fey/Lenssen/2019} and Torchmeta \citep{deleu2019torchmeta}. For all models the number and structure of the layers is as described in Section 4.2 of the paper, where we use 256-dimensional node embeddings at every layer.

At every cross-validation fold, 9 folds are used for training, and 1 for testing. Out of the 9 training folds, one is used as validation set. For each model we perform 100 iterations of hyperparameter optimization over the same search space (for shared parameters) using Ax \citep{Bakshy2018AEA}. 

We tried some sophisticated methods to balance the contribution of loss functions during multi-task training like GradNorm \citep{chen2018gradnorm} and Uncertainty Weights \citep{8578879}, but we saw that usually they do not positively impact performance. 
As an example, a GNN model trained for GC and LP with the classical procedure and using GradNorm, achieves results on GC, NC, and LP, that are on average $0.5\%$ higher than the same model trained without GradNorm. However, a GNN model trained for NC and LP with the classical procedure and using GradNorm achieves results that are $0.8\%$ lower than the same model trained without GradNorm. A similar behaviour happens by applying Uncertainty Weights, and, when the improvements were positive, they would never be higher than $1.7\%$.
\cite{v2020revisiting} also observe that when the multiple losses are of the same form, e.g. all cross-entropies, then these techniques tend to not bring any performance improvements. Furthermore, in the few cases where they increase performance, they work for both classically trained models, and for models trained with our proposed procedures. We then set the balancing weights to $\lambda^{(GC)} = \lambda^{(NC)} = \lambda^{(LP)} = 1$ to provide better comparisons between the training strategies. 

The experiments were run on a Nvidia 1080Ti GPU, and on a CPU cluster equipped with 8 cpus 12-Core Intel Xeon Gold 5118 @2.30GHz, with 1.5Tb of RAM.

\paragraph{Linear Model.} The linear model trained on the embeddings produced by our proposed method is a standard linear SVM. In particular we use the implementation available in Scikit-learn \citep{scikit-learn} with default hyperparameters. For graph classification, we take the mean of the node embeddings as input. For link prediction we take the concatenation of the embeddings of two nodes. For node classification we keep the embeddings unaltered.
\paragraph{Deep Learning Baselines.} We train the single task models for 1000 epochs, and the multi-task models for 5000 epochs, with early stopping on the validation set (for multi-task models we use the sum of the task validation losses or accuracies as metrics for early-stopping). Optimization is done using Adam \citep{Kingma2015AdamAM}. For node classification and link prediction we found that normalizing the node embeddings to unit norm in between GCN layers helps performance.
\paragraph{Our Meta-Learning Procedure.} We train the single task models for 5000 epochs, and the multi-task models for 15000 epochs, with early stopping on the validation set (for multi-task models we use the sum of the task validation losses or accuracies as metrics for early-stopping). Early stopping is very important in this case as it is the only way to check if the meta-learned model is overfitting the training data.
The inner loop adaptation consists of 1 step of gradient descent. Optimization in the outer loop is done using Adam \citep{Kingma2015AdamAM}. We found that normalizing the node embeddings to unit norm in between GCN layers helps performance.

\begin{table}[t]
  \caption{Results of a neural network trained on the embeddings generated by a multi-task model, to perform a task that was not seen during training by the multi-task model. ``$x$,$y$ -\textgreater $z$'' indicates that the multi-task model was trained on tasks $x$ and $y$, and the neural network is performing task $z$.}\label{tableQ2}
  \centering
  \begin{tabular}{llcccc}
    \hline
    \textbf{Task} & \textbf{Model} & \multicolumn{4}{c}{\textbf{Dataset}} \\
            &            & ENZYMES & PROTEINS & DHFR & COX2 \\
    \hline
    \multirow{3}{*}{GC,NC -\textgreater LP} & Cl & $56.9 \pm 3.9$ & $54.4 \pm 1.4$ & $61.2 \pm 2.2$ & $59.8 \pm 0.4$\\
                                                                   & iSAME & $77.3 \pm 4.5$ & $88.5 \pm 1.8$ & $99.8 \pm 1.8$ & $97.1 \pm 2.0$\\
                                                                   & eSAME & $78.9 \pm 2.8$ & $89.1 \pm 1.5$ & $99.7 \pm 2.2$ & $95.8 \pm 3.3$\\
    \hline
    \multirow{3}{*}{GC,LP -\textgreater NC} & Cl & $69.1 \pm 1.2$ & $57.3 \pm 1.6$ & $58.3 \pm 9.3$ & $68.9 \pm 10.7$\\
                                                                   & iSAME & $73.3 \pm 2.1$ & $59.2 \pm 2.5$ & $77.6 \pm 1.6$ & $78.1 \pm 4.6$\\
                                                                   & eSAME & $79.1 \pm 1.7$ & $64.7 \pm 3.0$ & $76.1 \pm 2.7$ & $76.9 \pm 3.3$\\
    \hline
    \multirow{3}{*}{NC,LP -\textgreater GC} & Cl & $47.1 \pm 2.4$ & $75.3 \pm 1.5$ & $77.5 \pm 3.1$ & $79.9 \pm 3.4$\\
                                                                   & iSAME & $48.5 \pm 5.5$ & $76.1 \pm 2.3$ & $76.1 \pm 3.7$ & $79.7 \pm 5.1$\\
                                                                   & eSAME & $56.6 \pm 3.1$ & $74.6 \pm 2.7$ & $77.1 \pm 3.6$ & $79.3 \pm 6.2$\\
    \hline
  \end{tabular}
\end{table}

 \section{Full Results for the Generalization of Node Embeddings}\label{fullresq2}
 Table \ref{tableQ2} contains results for a neural network, trained on the embeddings generated by a multi-task model, to perform a task that was not seen during the training of the multi-task model.
 Accuracy (\%) is used for node classification (NC) and graph classification (GC); ROC AUC (\%) is used for link prediction (LP). The embeddings produced by a model trained with SAME lead to higher performance (up to \textbf{35\%}), showing that our procedures lead to models that can generate more informative node embeddings with respect to the classical end-to-end training procedure.

\section{Ablation Study - Full Results}\label{fullresappendix}
\begin{table}
  \caption{Results for a single-task model trained in a classical supervised manner (Cl), and a \textbf{linear} classifier trained on the embeddings generated by a model trained with an ablated ``single-task'' version our meta-learning strategies ((a)iSAME, (a)eSAME).}\label{tablesingletask} 
  \centering
  \begin{tabular}{llcccc}
    \hline
    \textbf{Task} & \textbf{Model} & \multicolumn{4}{c}{\textbf{Dataset}} \\
            &            & ENZYMES & PROTEINS & DHFR & COX2 \\
    \hline
    \multirow{3}{*}{NC} & Cl & $87.5 \pm 1.9$ & $72.3 \pm 4.4$ & $97.3 \pm 0.2$ & $96.4 \pm 0.3$\\
                                    & (a)iSAME & $87.3 \pm 0.8$ & $81.8 \pm 1.6$ & $96.6 \pm 0.3$ & $96.1 \pm 0.4$\\
                                    & (a)eSAME & $87.8 \pm 0.7$ & $82.4 \pm 1.6$ & $96.8 \pm 0.2$ & $96.5 \pm 0.6$\\
    \hline
    \multirow{3}{*}{GC} & Cl & $51.6 \pm 4.2$ & $73.3 \pm 3.6$ & $71.5 \pm 2.3$ & $76.7 \pm 4.7$\\
                                   & (a)iSAME & $50.8 \pm 2.9$ & $73.5 \pm 1.2$ & $73.2 \pm 3.2$ & $76.3 \pm 4.6$\\
                                   & (a)eSAME & $52.1 \pm 5.0$ & $72.6 \pm 1.6$ & $71.6 \pm 2.4$ & $75.6 \pm 4.1$\\
    \hline
    \multirow{3}{*}{LP} & Cl & $75.5 \pm 3.0$ & $85.6 \pm 0.8$ & $98.8 \pm 0.7$ & $98.3 \pm 0.8$\\
                                   & (a)iSAME& $81.7 \pm 1.7$ & $84.0 \pm 1.1$ & $99.2 \pm 0.4$ & $99.1 \pm 0.5$\\
                                   & (a)eSAME & $80.1 \pm 3.4$ & $84.1 \pm 0.9$ & $99.2 \pm 0.3$ & $99.2 \pm 0.7$\\
    \hline
  \end{tabular}
\end{table}
Table \ref{tablesingletask} shows the results for the ablated ``single-task'' versions of SAME (i.e., the full results related to Figure 4 of the main paper).
For every task, we train a \textbf{linear classifier} on top of the embeddings produced by a model trained using our proposed methods, and compare against a network with the same architecture trained in a classical manner. For all three tasks, a \textbf{linear} classifier on the embeddings produced by a model trained with our methods achieves comparable, if not superior, performance to an end-to-end model. In fact, the linear classifier is never outperformed by more than 2\%, and it can outperform the classical end-to-end model by up to 12\%. These results highlight the impact of designing support and target sets to encourage generalization.

\begin{table*}[t]
  \caption{Results of a neural network trained on the embeddings generated by a multi-task model, to perform a task that was not seen during training by the multi-task model. The multi-task model has been trained with an ablated version of iSAME and eSAME (which we refer to as (a)iSAME and (a)eSAME), where no single-task adaptation is performed, but a multi-task version of the traditional meta-learning procedure is applied. ``$x$,$y$ -\textgreater $z$'' indicates that the multi-task model was trained on tasks $x$ and $y$, and the neural network is performing task $z$.}\label{traditionalMAMLonunseentasks}
  \centering
  \begin{tabular}{llcccc}
    \hline
    \textbf{Task} & \textbf{Model} & \multicolumn{4}{c}{\textbf{Dataset}} \\
            &            & ENZYMES & PROTEINS & DHFR & COX2 \\
    \hline
    \multirow{2}{*}{GC,NC -\textgreater LP} & (a)iSAME & $75.6 \pm 3.3$ & $88.3 \pm 1.2$ & $98.4 \pm 0.9$ & $95.1 \pm 1.7$\\
                                                                   & (a)eSAME & $79.4 \pm 2.8$ & $89.2 \pm 1.6$ & $97.4 \pm 0.7$ & $95.3 \pm 1.4$\\
    \hline
    \multirow{2}{*}{GC,LP -\textgreater NC} & (a)iSAME & $71.8 \pm 2.5$ & $59.7 \pm 3.2$ & $76.4 \pm 2.3$ & $79.1 \pm 2.3$\\
                                                                   & (a)eSAME & $79.5 \pm 1.6$ & $63.8 \pm 2.1$ & $77.0 \pm 2.1$ & $78.9 \pm 2.1$\\
    \hline
    \multirow{2}{*}{NC,LP -\textgreater GC} & (a)iSAME & $42.3 \pm 5.5$ & $75.8 \pm 2.6$ & $76.9 \pm 4.4$ & $78.3 \pm 7.5$\\
                                                                   & (a)eSAME & $53.6 \pm 3.5$ & $75.6 \pm 2.1$ & $77.3 \pm 2.8$ & $77.7 \pm 5.2$\\
    \hline
  \end{tabular}  
\end{table*}

\begin{table*}[t]
  \caption{$\Delta_m$ (\%) results for a \textbf{linear} classifier trained on the node embeddings generated by a model trained with an ablated version of iSAME and eSAME (which we refer to as (a)iSAME and (a)eSAME), where no single-task adaptation is performed, but a multi-task version of the traditional meta-learning procedure is applied.}\label{multitasktraditional}
  \centering
  \begin{tabular}{cccccccc}
    \hline
    \multicolumn{3}{c}{\textbf{Task}} & \textbf{Model} & \multicolumn{4}{c}{\textbf{Dataset}} \\
               GC & NC & LP     &            & ENZYMES & PROTEINS & DHFR & COX2 \\
    \hline
    \multirow{2}{*}{\checkmark} & \multirow{2}{*}{\checkmark} & \multirow{2}{*}{ } & (a)iSAME & $-3.2 \pm 0.4$ & $2.8 \pm 1.3$ & $-0.4 \pm 0.3$ & $-0.5 \pm 0.2$\\
                                                     &&                                                                    & (a)eSAME & $0.5 \pm 1.1$ & $1.7 \pm 0.5$ & $-0.5 \pm 1.2$ & $-1.5 \pm 0.5$\\
    \hline
    \multirow{2}{*}{\checkmark} & \multirow{2}{*}{ } & \multirow{2}{*}{\checkmark} & (a)iSAME & $1.5 \pm 2.9$ & $-0.9 \pm 1.1$ & $0.1 \pm 4.1$ & $-0.2 \pm 3.1$\\
                                                         &&                                                                    & (a)eSAME & $3.0 \pm 1.7$ & $-2.5 \pm 1.6$ & $-0.4 \pm 3.5$ & $-0.4 \pm 4.2$\\
    \hline
    \multirow{2}{*}{ } & \multirow{2}{*}{\checkmark} & \multirow{2}{*}{\checkmark} & (a)iSAME & $4.3 \pm 2.8$ & $4.5 \pm 1.1$ & $-0.2 \pm 6.2$ & $-0.6 \pm 4.2$\\
                                                         &&                                                                    & (a)eSAME & $3.1 \pm 0.4$ & $5.0 \pm 1.1$ & $-0.1 \pm 5.9$ & $-0.5 \pm 4.1$\\
    \hline
    \multirow{2}{*}{\checkmark} & \multirow{2}{*}{\checkmark} & \multirow{2}{*}{\checkmark} & (a)iSAME  & $ 1.0 \pm 1.2$ & $2.7 \pm 0.3$ & $-1.2 \pm 2.4$ & $-1.1 \pm 2.9$\\
                                                     &&                                                                    & (a)eSAME & $0.4 \pm 1.2$ & $1.1 \pm 0.3$ & $-1.3 \pm 1.1$ & $-0.9 \pm 1.1$\\
    \hline
  \end{tabular}
\end{table*}

Table \ref{traditionalMAMLonunseentasks} presents the results of a neural network trained on the embeddings generated by a model trained with an ablated version of iSAME and eSAME to perform a task that was unseen during training. The ablated version of iSAME and eSAME is obtained by using using the support and target set design that encourage generalization, but by not applying the multiple separate single-task adaptations in the inner loop. In particular all tasks are performed concurrently on all graphs both in the inner loop and the outer loop (leading to a \textit{multi-task} version of the traditional MAML and ANIL procedure, but with our design of support and target sets). 
Finally, Table \ref{multitasktraditional} shows the results in terms of the multi-task performance ($\Delta_m$) metric \citep{8954118} of the latter ablated version of SAME. 
For both experiments, we notice that the results are not significantly different from those of the non-ablated iSAME and eSAME (Table \ref{tableQ2} in Appendix, and Table 2 of the paper), indicating that iSAME and eSAME increase the sample efficiency of the training procedure as they can lead to a model generating embeddings that can reach the same results but by only applying one task per graph (and hence only requiring the labels for one task per graph).
\end{document}